\documentclass{article} % For LaTeX2e
\usepackage{iclr2027_conference,times}

\usepackage{amsmath,amsfonts,bm}

\def\eqref#1{equation~\ref{#1}}
\def\1{\bm{1}}

\DeclareMathAlphabet{\mathsfit}{\encodingdefault}{\sfdefault}{m}{sl}
\SetMathAlphabet{\mathsfit}{bold}{\encodingdefault}{\sfdefault}{bx}{n}

\usepackage{hyperref}
\usepackage{url}
\usepackage{graphicx}
\usepackage{booktabs}
\usepackage{capt-of}
\usepackage{wrapfig}
\usepackage{booktabs,multirow,amssymb,array,colortbl,graphicx}
\usepackage{xcolor}
\usepackage{nicematrix}
\usepackage{booktabs,multirow,amssymb}
\definecolor{bestbg}{RGB}{242,241,247}
\usepackage{graphicx}
\usepackage{subcaption}
\usepackage{enumitem}
\setlist[itemize]{noitemsep, nolistsep, leftmargin=*}
\title{DexForge: High-Fidelity Physics-Informed\\
Dexterous Retargeting}

\author{
Meizhong Wang\textsuperscript{1},
Kun Cao\textsuperscript{1,2}\thanks{Corresponding author: Kun Cao, email: \texttt{caokun@tongji.edu.cn}},
Ruiqi Ni\textsuperscript{3},
Lihua Xie\textsuperscript{4},
Yiguang Hong\textsuperscript{1,2}\\
\textsuperscript{1}Tongji University,
\textsuperscript{2}Shanghai Research Institute for Intelligent Autonomous Systems\\
\textsuperscript{3}Purdue University,
\textsuperscript{4}Nanyang Technological University, Singapore
}

\iclrfinalcopy
\begin{document}
\raggedbottom

\maketitle

\begin{abstract}
Human demonstrations offer rich examples of precise dexterous manipulation and a promising source of robot training data. However, high-fidelity reproduction of demonstrated motions and hand--object interactions across robot embodiments remains challenging under physical constraints. We present DexForge, a differentiable physics-grounded framework for converting human video demonstrations into high-fidelity robot trajectories. We reconstruct spherical-Gaussian object models and hand--object motion from visual observations, then build a differentiable simulator combining efficient Gaussian collision detection with existing differentiable dynamics. Based on this simulator, DexForge combines contact-aware kinematic retargeting with force-aware dynamics retargeting: robot-adapted stable contacts guide kinematic reference construction and subsequent gradient-based control refinement for precise physical motion reproduction. Experiments on 130 DexYCB and HOT3D demonstrations across seven dexterous hands show success-rate gains of approximately 35--53 percentage points over the baseline, with object position and orientation tracking errors on successful trajectories reduced by approximately 34--67\% and 71--78\%, respectively. Further experiments demonstrate open-loop transfer to MuJoCo and real-robot execution. Our project page is available at \url{https://wmz1226.github.io/DexForge/}.
\end{abstract}
\vspace{-10pt}
\section{Introduction}
\vspace{-10pt}
\label{sec:introduction}

Enabling robots with different embodiments to reproduce precise human manipulation is an important step toward transferring human expertise to robots~\citep{xu2025dexumi,zhang2026unidex}. High-fidelity reproduction requires more than task completion; the demonstrated motions and hand--object interactions must also be preserved. For example, establishing a stable grasp is only the starting point; precisely controlling the object's subsequent movement and orientation captures finer aspects of manipulation. Human manipulation videos contain rich examples of these behaviors, yet translating them into high-fidelity trajectories executable by different dexterous hands remains challenging.

Recent advances in learning from human demonstrations and retargeting are narrowing this gap. Hand--object reconstruction recovers explicit 3D geometry and motion references from videos~\citep{wang2025magichoi,shi2026agile}, while kinematic retargeting adapts these references to robots, increasingly incorporating contact preservation~\citep{lakshmipathy2025kinematic} and collision constraints~\citep{tian2026kilohertz} to improve geometric consistency. Building on these references, physics-based retargeting uses policy learning~\citep{li2025maniptrans} or trajectory optimization~\citep{pan2026spiderscalablephysicsinformeddexterous} to find controls that drive object motion in simulation, with contact guidance helping preserve demonstrated interactions~\citep{mandi2025dexmachina}. Meanwhile, advances in differentiable collision detection~\citep{beker2025smooth,wang2026contactgaussian} and dynamics modeling~\citep{paulus2026differentiable} enable control optimization through simulation gradients, providing another route to precise trajectory refinement~\citep{liu2024parameterized}. Together, these developments extend retargeting beyond geometric hand-motion matching toward physically grounded manipulation reproduction.

Despite these advances, high-fidelity reproduction across different robot hands remains difficult. Several problems need to be solved in sequence. First, the reconstructed object geometry must be accurate enough for collision and contact reasoning. Second, contacts from the human demonstration may not be reachable or stable for a robot with a different hand shape. Third, even a penetration-free robot trajectory with geometrically appropriate contacts may not be executable under contact dynamics and actuation constraints. Physics-based optimization also involves a trade-off. Sampling-based methods can handle complex contact changes, but obtaining very accurate trajectories may require many samples. Gradient-based optimization can refine trajectories more directly, but it is sensitive to initialization and local minima. This motivates us to use robot-adapted contacts to provide a good kinematic reference and continued guidance during gradient-based dynamics optimization.

We introduce DexForge, a differentiable physics-grounded framework for high-fidelity retargeting from human video demonstrations. Our key insight is that human contacts should not be copied directly to a robot with a different hand. Instead, we first adapt these contacts to the target robot, and then use the adapted contacts to guide both kinematic retargeting and dynamic control optimization. To support this process, we reconstruct spherical Gaussian object models that provide both visual appearance for pose estimation and geometry for collision checking. We then extract stable contacts from the human demonstration and refine them together with robot configurations so that they are reachable, collision-free, and stable for the target hand. These contacts are used to build a robot kinematic reference. Finally, we combine differentiable Gaussian collision with ComFree dynamics and optimize robot controls through simulation gradients to reproduce the demonstrated motion more accurately. We further use system identification to reduce the mismatch between simulation and the target system. Our contributions are:

\begin{itemize}[leftmargin=*]
    \item We propose DexForge, a complete framework that combines visual reconstruction, contact-aware kinematic retargeting, and force-aware dynamics optimization to transform human video demonstrations into high-fidelity, executable demonstrations across robot embodiments.
    \item We introduce robot-adapted stable-contact guidance through joint anchor and robot-configuration optimization, supporting both kinematic reference construction and dynamic refinement.
    \item We jointly train spherical Gaussians for appearance and geometry and develop efficient differentiable collision detection for them, supporting integration with existing simulators.
    \item We evaluate DexForge on 130 DexYCB and HOT3D demonstrations across seven hands, outperforming sampling-based baselines in retargeting success and object-tracking accuracy, and demonstrate open-loop MuJoCo transfer and real-robot execution.
\end{itemize}

\vspace{-10pt}
\section{Related Work}
\vspace{-10pt}
\label{sec:related-work}
\noindent\textbf{Dexterous Manipulation from Human Demonstrations.} Human demonstrations can be collected through portable motion-capture systems~\citep{wang2024dexcapscalableportablemocap} and wearable interfaces~\citep{xu2025dexumi,tao2025dexwild}. Egocentric human videos also provide scalable supervision for dexterous policy learning~\citep{zhang2026unidex,zheng2026egoscalescalingdexterousmanipulation}. Explicit scene and motion references recovered from videos enable robot trajectory generation~\citep{mu2026deximit,paliwal2026idodexterousmanipulation} or guide policy learning in simulation~\citep{lum2025crossinghumanrobotembodimentgap}. DexForge reconstructs object models from separate capture sequences and recovers hand--object motion from demonstration videos, then converts these references into \emph{high-fidelity} robot trajectories through \emph{contact adaptation and physical optimization}.

\noindent\textbf{Dexterous Motion Retargeting.} Kinematic retargeting maps human motion to robot configurations through geometric correspondence objectives~\citep{yin2025geometricretargetingprincipledultrafast,xin2026analyzing}. Interaction-aware methods further preserve contact regions or local hand--object geometry~\citep{lakshmipathy2025kinematic,wu2026toporetarget}, while constrained formulations incorporate collision avoidance~\citep{tian2026kilohertz}. Such interaction structure can also be encoded through interaction meshes for whole-body retargeting~\citep{yang2026omniretargetinteractionpreservingdatageneration} and dexterous manipulation~\citep{feng2026minimalist}. The kinematic stage of C2Dex~\citep{ren2026c2dex} uses stable object-side contacts to guide reference construction. For grasp configurations, GenHand~\citep{qi2026genhand} refines human-derived contact anchors through force-closure optimization, while GraspQP~\citep{zurbrugg2025graspqp} develops a differentiable force-closure energy for grasp synthesis. DexForge \emph{jointly} optimizes robot keyframe \emph{configurations and anchors} shared within stable-contact intervals, considering \emph{reachability and force closure}.

Physics-based retargeting accounts for contact dynamics, with representative approaches using reference-guided reinforcement learning~\citep{li2025maniptrans,liu2025dextrack,mandi2025dexmachina,xu2025dexplore,ren2026c2dex,feng2026minimalist} or sampling-based trajectory optimization~\citep{yang2025physics,pan2026spiderscalablephysicsinformeddexterous,paliwal2026idodexterousmanipulation}. SPIDER~\citep{pan2026spiderscalablephysicsinformeddexterous} progressively relaxes virtual contact guidance during optimization, while Do as I Do~\citep{paliwal2026idodexterousmanipulation} adapts sampling-based retargeting to noisy reconstructed references. Gradient-based transfer is also explored by QuasiSim~\citep{liu2024parameterized}, which progressively tightens physical constraints through a curriculum of parameterized quasi-physical simulators. DexForge uses \emph{robot-adapted stable contacts} to construct collision-constrained kinematic references for initialization and retains the same refined anchors as interaction targets during \emph{gradient-based} dynamics optimization.

\noindent\textbf{Differentiable Simulation and Physical Scene Modeling.} Differentiable collision detection provides geometric derivatives for optimization~\citep{tracy2023differentiable,montaut2023differentiable,beker2026smoothly}, while contact models and simulators advance differentiable dynamics~\citep{jin2024complementarity,ye2026efficient,paulus2026differentiable}. Gaussian representations~\citep{kerbl20233d} also support physical modeling: ContactGaussian-WM~\citep{wang2026contactgaussian} unifies appearance and collision geometry with spherical Gaussians but optimizes geometry and appearance separately, whereas D-REX~\citep{lou2026drexdifferentiablerealtosimtorealengine} combines Gaussian rendering and mesh-based collisions with differentiable mass identification and grasping policy learning. DexForge \emph{jointly} optimizes spherical Gaussians to better capture appearance and collision geometry, providing an \emph{efficient differentiable collision-query interface} directly on these primitives. Integrating it with ComFree dynamics~\citep{borse2026comfree} yields ComFree-GS, supporting contact-aware kinematic retargeting and gradient-based physical retargeting.

\vspace{-10pt}
\section{Method}
\begin{figure}[htbp]
\centering
\includegraphics[width=\linewidth]{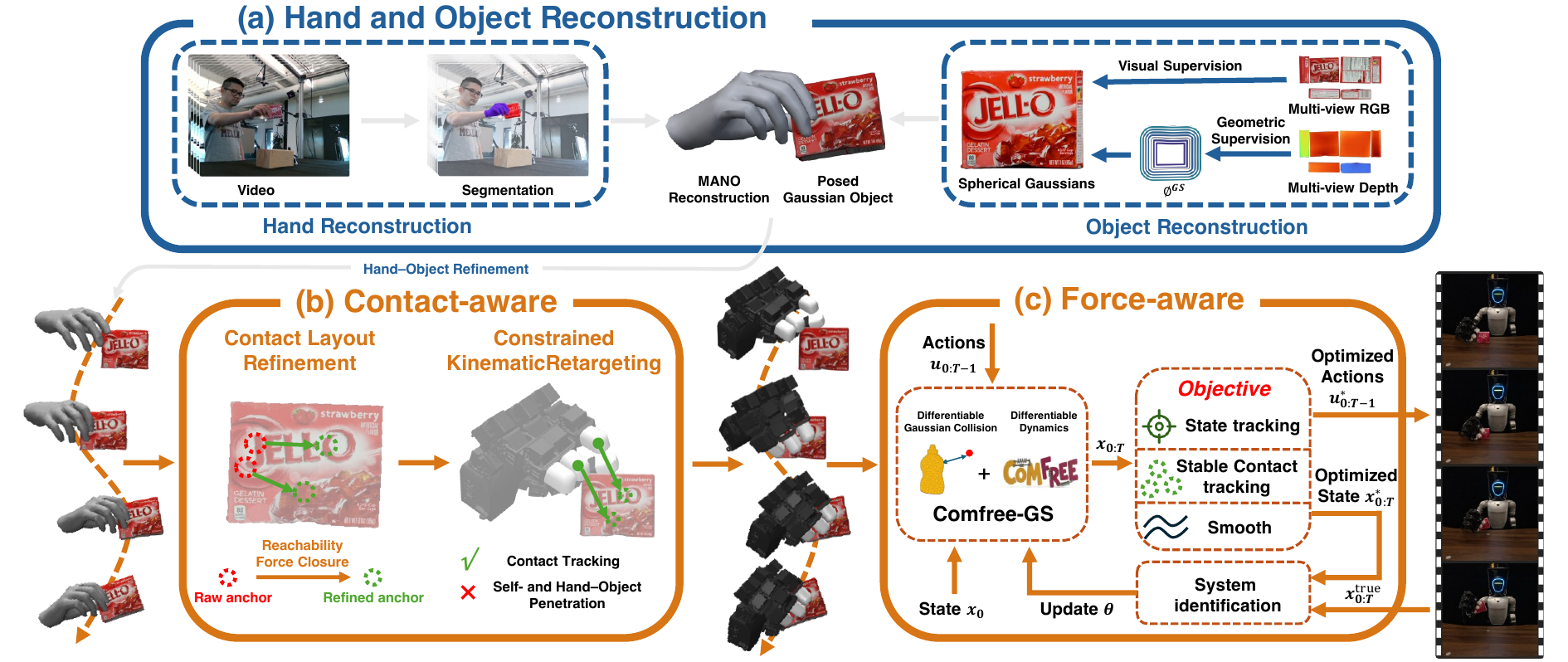}
\caption{Overview of DexForge. (a) Hand and object reconstruction builds spherical Gaussian object models under RGB-D supervision and recovers hand--object motion from demonstration videos. (b) Contact-aware kinematic retargeting refines demonstrated contacts for robot reachability and contact stability, generating penetration-free references that track them. (c) Force-aware dynamics retargeting optimizes controls through simulation gradients in ComFree-GS, which combines efficient differentiable Gaussian collision detection with ComFree dynamics. System identification further reduces target-system mismatch to support high-fidelity physical reproduction.}
\label{fig:pipeline}
\end{figure}

\vspace{-10pt}
Our goal is to transform human video demonstrations into high-fidelity, physically executable robot rollouts. We formulate the problem in Section~\ref{sec:Problem Formulation} and address it by progressively establishing the physical structure required for robot execution, as illustrated in Figure~\ref{fig:pipeline}. We first reconstruct the geometry of the hand and object and recover their pose trajectories from visual observations (Section~\ref{sec:object_reconstruction}). We then extract stable human contacts, adapt them to the target robot, and generate a robot-adapted kinematic reference that satisfies collision constraints (Section~\ref{sec:Kinematic Retargeting}). Finally, we realize this reference by optimizing controls through a differentiable simulator and calibrating the dynamics model against observations from the target system (Section~\ref{sec:Dynamics-aware retargeting}).

\vspace{-10pt}
\subsection{Problem Formulation}
\vspace{-5pt}
\label{sec:Problem Formulation}
Given a hand--object interaction video $V$, we recover geometric and motion references for subsequent retargeting. We construct a Gaussian object model $O$ from a separate multi-view RGB-D video and use it to recover the object pose trajectory $q_{0:T}^{\mathrm{ref,object}}$ in $V$. MANO~\citep{romero2022embodied} hand reconstruction from $V$ provides the keypoint trajectory $\mathbf{P}_{0:T}^{\mathrm{mano}}$. Here, $\mathbf{P}_t^{\mathrm{mano}}\in\mathbb{R}^{N_k\times 3}$ contains the positions of $N_k$ hand keypoints, and $q_t^{\mathrm{ref,object}}\in\mathrm{SE}(3)$ denotes the object pose. Both trajectories are expressed in a common coordinate frame to support subsequent retargeting.

Given these geometric and motion references, we formulate kinematic and dynamics retargeting as two sequential constrained optimization problems with complementary roles. \textbf{Kinematic retargeting} seeks a robot reference trajectory $q_{0:T}^{\mathrm{ref,robot}}$ that matches the demonstrated motion through keypoint fitting, subject to configuration and collision constraints. Specifically, we use the correspondence set $\mathcal{C}$ to fit robot keypoint positions $\mathbf{P}_i^{\mathrm{robot}}(q_t^{\mathrm{ref,robot}})$, computed by forward kinematics, to their MANO counterparts $\mathbf{P}_{t,j}^{\mathrm{mano}}$, where $(i,j)\in\mathcal{C}$. The robot and object reference trajectories define $q_t^{\mathrm{ref}}=(q_t^{\mathrm{ref,robot}},q_t^{\mathrm{ref,object}})$ and the reference state $x_t^{\mathrm{ref}}=(q_t^{\mathrm{ref}},\dot{q}_t^{\mathrm{ref}})$. \textbf{Dynamics retargeting} then optimizes robot controls $\mathbf{u}_{0:T-1}$ to track the reference state trajectory $x_{0:T}^{\mathrm{ref}}$, starting from $x_0=x_0^{\mathrm{ref}}$. The resulting trajectory must satisfy the simulated dynamics $x_{t+1}=f(x_t,\mathbf{u}_t,t)$ and the admissible state and control constraints $x_t\in\mathcal{X}$ and $\mathbf{u}_t\in\mathcal{U}$. Here, $x_t=(q_t,\dot{q}_t)$ denotes the physical state, with $q_t=(q_t^{\mathrm{robot}},q_t^{\mathrm{object}})$. To reduce mismatch with the target system, we calibrate the simulator's physical parameters $\theta$ through system identification, minimizing the discrepancy between simulated trajectories $x(\theta)$ and observed trajectories $x^{\mathrm{target}}$ under identical controls.
\vspace{-10pt}
\subsection{Hand and Object Reconstruction}
\vspace{-5pt}
\label{sec:object_reconstruction}
In this section, we reconstruct high-fidelity object assets from multi-view RGB-D videos and recover object pose trajectories and MANO hand motion from hand--object interaction videos. The reconstructed assets must combine accurate geometry for physical simulation in subsequent retargeting with faithful appearance for pose estimation. ContactGaussian-WM~\citep{wang2026contactgaussian} demonstrates how spherical Gaussians can unify visual appearance and collision geometry, avoiding the need to maintain and align separate models. However, its image-based initialization does not directly constrain interior distances, potentially yielding unreliable penetration-depth estimates. Its subsequent color-fitting stage holds this potentially biased geometry fixed, which can also limit appearance reconstruction accuracy. We therefore introduce depth-derived distance supervision to constrain geometry while allowing geometry and color to be optimized jointly. Specifically, we define the Gaussian collision distance field at a query point $p_i\in\mathbb{R}^3$ as:
\begin{equation}
\phi^{\mathrm{GS}}(p_i)
=
\min_j
\left(
\left\|p_i-\boldsymbol{\mu}_j\right\|_2-r_j
\right),
\quad 1\leq j\leq N_{\mathrm{gs}}.
\label{eq:gs_sdf_single}
\end{equation}
Here, $N_{\mathrm{gs}}$ is the number of spherical Gaussian primitives, $\boldsymbol{\mu}_j$ and $\sigma_j$ are the center and scale of the $j$-th Gaussian, and $r_j=3\sigma_j$ is its empirically chosen collision radius. We construct a reference signed distance field from multi-view depth maps and use it to supervise $\phi^{\mathrm{GS}}(p_i)$, combining the distance-matching loss with the standard 3DGS objective to jointly optimize geometry and color.

We use GS-Pose~\citep{cai2024gsposegeneralizablesegmentationbased6d} with the reconstructed Gaussian object model $O$ to recover its pose trajectory $q^{\mathrm{ref,object}}_{0:T}$ in the interaction video $V$. Keeping the model's geometry and appearance fixed, pose refinement optimizes the object's rotation and translation through differentiable rendering to align its rendered appearance with the observations. We then also recover MANO hand motion from $V$ using HaMeR~\citep{pavlakos2023reconstructinghands3dtransformers} predictions and camera calibration, obtaining the keypoint trajectory $\mathbf{P}^{\mathrm{mano}}_{0:T}$ in the same coordinate frame as the object poses.

To support collision simulation for retargeting, we develop an efficient differentiable interface for the reconstructed Gaussians, first extending Eq.~(\ref{eq:gs_sdf_single}) to the general Gaussian--Gaussian form:
\begin{equation}
\phi^{\mathrm{GS}}(\mathcal{A},\mathcal{B})
=
\min_{i,j}
\left(
\left\|\boldsymbol{\mu}^{\mathcal{A}}_i-\boldsymbol{\mu}^{\mathcal{B}}_j\right\|_2
-r^{\mathcal{A}}_i-r^{\mathcal{B}}_j
\right),
\quad
1\leq i\leq N^{\mathcal{A}},\;
1\leq j\leq N^{\mathcal{B}}.
\label{eq:gs_sdf_pair}
\end{equation}
Here, $\mathcal{A}$ and $\mathcal{B}$ contain $N^{\mathcal{A}}$ and $N^{\mathcal{B}}$ spherical Gaussians, respectively; $\boldsymbol{\mu}^{\mathcal{A}}_i$ and $r^{\mathcal{A}}_i$ denote the center and collision radius of Gaussian $i$ in $\mathcal{A}$. All centers share a common coordinate frame. We implement efficient differentiable collision backends separately in JAX and Warp. Gaussian training, motion reconstruction and collision detection are detailed in Appendices~\ref{app:gaussian-training}, ~\ref{app:motion-reconstruction}  and \ref{app:bvh}, respectively.

\vspace{-10pt}
\subsection{Contact-Aware Kinematic Retargeting}
\vspace{-5pt}
\label{sec:Kinematic Retargeting}
The hand--object trajectories recovered in Section~\ref{sec:object_reconstruction} describe the demonstrated interaction but do not directly specify how a robot with a different morphology should reproduce it. In particular, morphological differences can make human contacts unreachable or unsuitable, while MANO reconstruction errors, such as substantial hand--object penetration, can make extracted anchors unreliable. We therefore extract stable contacts and adapt them to the target embodiment while preserving the demonstrated interaction semantics. These adapted contacts then guide full-trajectory optimization under configuration and collision constraints, yielding a robot-compatible kinematic reference without yet assuming that the motion is dynamically executable.

\noindent\textbf{Robot-Adapted Stable-Contact Guidance.} We first extract raw stable contacts by evaluating the distance between each MANO fingertip surface point $j$ and the object Gaussians at each frame $t$ using $\phi^{\mathrm{GS}}(O(q_t^{\mathrm{ref,object}}),\mathbf{P}^{\mathrm{mano}}_{t,j})$ in Eq.~(\ref{eq:gs_sdf_pair}). We summarize stable contacts as triplets $(j,\mathbf{a}_{j,\mathcal{I}},\mathcal{I})$, each describing persistent contact between fingertip point $j$ and an object-frame anchor $\mathbf{a}_{j,\mathcal{I}}\in\mathbb{R}^3$ over an interval $\mathcal{I}=[t_1,t_2]$. Using the correspondence set $\mathcal{C}$, we associate each MANO contact point $j$ with robot keypoint $i$ and denote its corresponding anchor by $\mathbf{a}_{i,\mathcal{I}}$. To adapt these contacts to the target hand, we select $m$ keyframes at timesteps $\{t_k\}_{k\in\mathcal{K}}$ based on the extracted stable contacts and jointly optimize their robot configurations and contact anchors. Each anchor is constrained to the object surface and shared across keyframes within its contact interval:
\begin{subequations}
\label{eq:contact-refinement}
\begin{align}
\min_{\{q_{t_k}^{\mathrm{key,robot}}\},\,\{\hat{\mathbf{a}}_{i,\mathcal{I}}\}}\quad
&\sum_{k\in\mathcal{K}}
\sum_{(i,\mathcal{I}):\,t_k\in\mathcal{I}}
\left\|
\mathbf{P}^{\mathrm{robot}}_i(q_{t_k}^{\mathrm{key,robot}})
-q_{t_k}^{\mathrm{ref,object}}\hat{\mathbf{a}}_{i,\mathcal{I}}
\right\|_2^2
+\mathcal{L}_{\mathrm{a}}
+\mathcal{L}_{\mathrm{fc}}
+\mathcal{L}_{\mathrm{reg}}
\label{eq:contact-refinement-objective}
\\
\mathrm{s.t.}\quad
&q_{t_k}^{\mathrm{key,robot}}\in\mathcal{Q}^{R},
\qquad \Phi_{t_k}\geq 0,
\qquad \phi^{\mathrm{GS}}(O,\hat{\mathbf{a}}_{i,\mathcal{I}})=0.
\label{eq:contact-refinement-feasibility}
\end{align}
\end{subequations}
The first term, the contact-fitting loss $\mathcal{L}_{\mathrm{c}}$, fits robot contact points to their shared refined anchors at keyframes within the corresponding contact intervals. The anchor term $\mathcal{L}_{\mathrm{a}}$ penalizes deviations of the refined anchor layout from the original layout. The force-closure term $\mathcal{L}_{\mathrm{fc}}$ aggregates losses over keyframes, penalizing the net wrench of bounded friction-cone contact forces and insufficient force-closure margins. The regularization term $\mathcal{L}_{\mathrm{reg}}$ tracks the MANO palm pose and hand shape, aligns contact normals, and smooths motion across keyframes. Here, $\Phi_t$ collects hand--object distances from Eq.~(\ref{eq:gs_sdf_pair}) and finger self-collision distances between link capsules. The constraints enforce configuration limits, nonpenetration, and anchors on the Gaussian surface. We use sequential quadratic programming (SQP)~\citep{bonnans2006numerical} with elastic relaxation. The refined anchors guide both retargeting stages. Loss definitions and solver settings are provided in Appendix~\ref{app:contact-refinement}.

\noindent\textbf{Constrained Kinematic Retargeting.} Using the refined anchors as contact targets throughout their corresponding contact intervals, we next optimize the full kinematic trajectory while fixing collision-feasible key configurations indexed by $\mathcal{K}_{\mathrm{f}}\subseteq\mathcal{K}$ and enforcing per-frame collision constraints:
\begin{subequations}
\label{eq:contact_qp}
\begin{align}
\min_{\{q_t^{\mathrm{ref,robot}}\}}\quad
&\sum_{t=0}^{T}\sum_{(i,j)\in\mathcal{C}}
w_{t,i}\big\|\mathbf{P}^{\mathrm{robot}}_i(q_t^{\mathrm{ref,robot}})-\mathbf{v}_{t,i}\big\|_2^2
+\mathcal{L}_{\mathrm{reg}}
\\
\mathrm{s.t.}\quad
&q_{t_k}^{\mathrm{ref,robot}}=q_{t_k}^{\mathrm{key,robot}},
\quad k\in\mathcal{K}_{\mathrm{f}},
\qquad \Phi_t\geq 0,
\qquad q_t^{\mathrm{ref,robot}}\in\mathcal{Q}^{R},
\\
\mathrm{with}\quad
&\mathbf{v}_{t,i}
=(1-\alpha_{t,i})\mathbf{P}^{\mathrm{mano}}_{t,j}
+\alpha_{t,i}q_t^{\mathrm{ref,object}}\hat{\mathbf{a}}_{i,\mathcal{I}}.
\end{align}
\end{subequations}
Here, $w_{t,i}$ weights position fitting, and $\mathbf{P}^{\mathrm{robot}}_i$ denotes the position of the mapped contact query point, or the fingertip site ($\alpha_{t,i}\equiv0$) for fingers without stable contacts. $\alpha_{t,i}\in\{0,1\}$ switches the target from the MANO correspondence to the refined contact anchor during stable contact. $\mathcal{L}_{\mathrm{reg}}$ tracks the MANO hand shape and palm pose, aligns contact normals, and penalizes velocity and acceleration over the full trajectory. The palm reference is adjusted by interpolated keyframe offsets. Starting from a trajectory repaired to be collision-free at every frame, we solve the problem with Gauss--Newton iterations, each linearizing the residuals and constraints and solving a trust-region QP over all frames; a line search maintains collision feasibility at every frame (Appendix~\ref{app:contact-refinement}).

\vspace{-10pt}
\subsection{Force-Aware Dynamics Retargeting}
\vspace{-5pt}
\label{sec:Dynamics-aware retargeting}
Kinematic retargeting provides a reference motion but does not determine the controls and contact forces required for execution. Force-aware dynamics retargeting therefore optimizes controls through a differentiable simulator, accounting for hand--object contact forces to generate trajectories under the modeled actuation and contact physics. We further calibrate the simulator's physical parameters to reduce model mismatch and support transfer to the target system.

\noindent\textbf{Force-Aware Control Optimization.} We integrate the Gaussian collision interface from Section~\ref{sec:object_reconstruction} into ComFree-Sim~\citep{borse2026comfree} to obtain our differentiable simulator, ComFree-GS. We use multi-start gradient-based MPC, with gradients backpropagated through ComFree-GS. At each replanning step $t$, we optimize position-control knots $\mathbf{U}$ over an $L$-step rollout horizon:
\begin{equation}
\label{eq:mpc}
\begin{aligned}
\min_{\mathbf{U}}\quad
& \|\mathbf{e}_{t+L}\|_{\mathbf{Q}_{\mathrm{f}}}^{2}
+\sum_{m=t+1}^{t+L-1}\|\mathbf{e}_m\|_{\mathbf{Q}}^{2}
+J_{\mathrm{c}}(\mathbf{x})
+J_{\mathrm{col}}(\mathbf{x})
+\mathcal{R}(\mathbf{U},\mathbf{x}) \\
\mathrm{s.t.}\quad
& x_{m+1}=f_{\theta}(x_m,\mathbf{u}_m),
\quad \mathbf{u}_m=[\operatorname{Interp}(\mathbf{U})]_m,
\quad m=t,\ldots,t+L-1.
\end{aligned}
\end{equation}
Here, $\mathbf{e}_m$ measures robot and object configuration errors, and $\operatorname{Interp}$ interpolates control knots into per-step commands. $J_{\mathrm{c}}$ combines weighted refined-anchor alignment with contact-depth penalties; $J_{\mathrm{col}}$ penalizes excessive penetration and inter-finger collisions. $\mathbf{Q}$ and $\mathbf{Q}_{\mathrm{f}}$ weight intermediate and terminal tracking errors, while $\mathcal{R}$ regularizes control differences and hand velocities and accelerations. Contact forces are computed by the simulator. The current state is fixed, allowing only a bounded initial hand-pose correction in the first window. We optimize $N$ candidates in parallel, keeping one warm start and perturbing the others, and select the lowest-cost numerically valid solution. We execute its initial portion, shift the horizon, and replan to generate optimized robot state and control trajectories. Implementation details are provided in Appendix~\ref{app:forceaware}.

\noindent\textbf{System Identification.}
\label{sec:system identification}
To reduce model mismatch during transfer to the target system $f_{\mathrm{target}}$, we calibrate the object and contact parameters of our simulator $f$. We replay controls $\mathbf{u}_{\mathrm{stable}}$ from stable-contact segments open-loop in $f_{\mathrm{target}}$ to obtain target trajectories $x^{\mathrm{target}}_{\mathrm{stable}}$. Starting each simulated segment from its observed initial state $x_0=x_0^{\mathrm{target}}$, we identify $\theta$ by minimizing $\|x_{\mathrm{stable}}(\theta)-x_{\mathrm{stable}}^{\mathrm{target}}\|_2^2$, with dynamics $x_{k+1}=f(x_k,\mathbf{u}_k;\theta)$ under the same replayed controls. Here, $\theta=(m,\mathbf{I},\mu_c,\delta)$ comprises the object mass, inertia tensor, contact friction coefficient, and collision-distance offset. The offset adjusts the effective collision surface to compensate for residual geometric error. Parameter coupling may yield a multimodal identification landscape, making gradient-only optimization sensitive to local minima and sampling-only search costly. We therefore use TSMC~\citep{yang2026tempered} to combine particle-based exploration with gradient-informed refinement. The calibrated parameters update the simulator for subsequent dynamics retargeting.

\vspace{-10pt}
\section{Experiments}
\vspace{-10pt}
This section presents extensive experiments to address three questions: 
(1) \textbf{Gaussian Representation Quality.} Can our spherical Gaussians better capture object geometry and appearance? (2) \textbf{Trajectory Fidelity and Feasibility.} Can our framework generate high-fidelity, physically executable robot trajectories in simulation, and how do individual components contribute? (3) \textbf{Trajectory Transferability.} Can the generated trajectories transfer to other simulators and real robots?

\vspace{-10pt}
\subsection{Q1: Gaussian Representation Quality}
\vspace{-5pt}
\begin{wrapfigure}{r}{0.4\textwidth}
\vspace{\dimexpr-\intextsep+5pt\relax}
\centering
\resizebox{\linewidth}{!}{%
\begin{tabular}{lcc}
\toprule
Method & PSNR (dB) $\uparrow$ & RMSE (mm) $\downarrow$ \\
\midrule
3DGS (spherical) & \textbf{45.600} & 2.5705 \\
ContactGaussian-WM & 42.397 & 1.4979 \\
Ours & 43.564 & \textbf{0.4065} \\
\bottomrule
\end{tabular}%
}
\par\smallskip
\includegraphics[width=\linewidth]{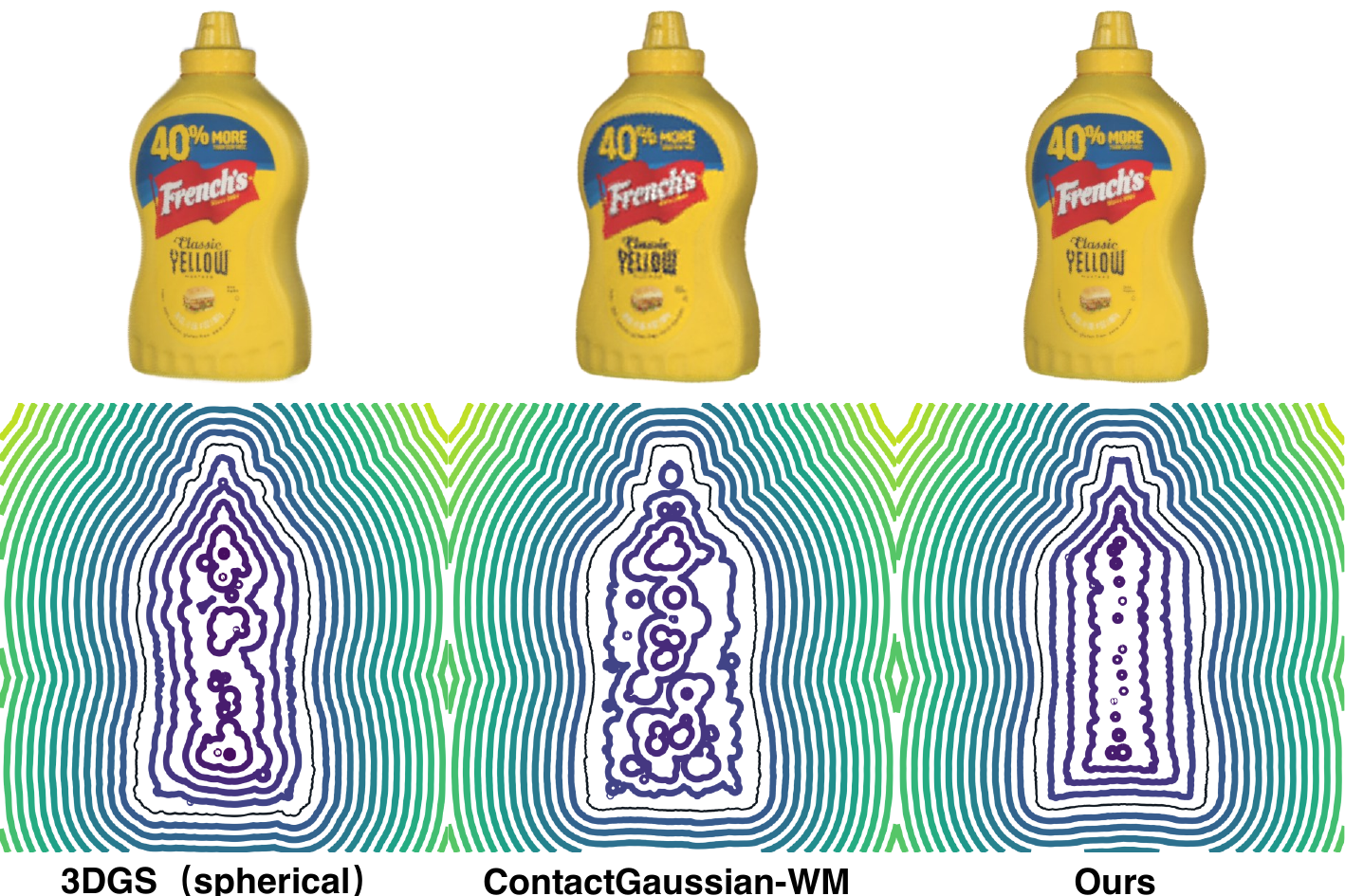}
\caption{Quantitative results on all 20 objects in DexYCB (top); front-view rendering and distance-field visualization for a representative object (bottom).}
\label{fig:gaussian-quality}
\vspace{-10pt}
\end{wrapfigure}
We evaluate the rendering quality and geometric accuracy of our Gaussian representation on all 20 objects in DexYCB~\citep{chao2021dexycb}. For each object, we render RGB-D images from 72 viewpoints. We compare against a spherical variant of standard 3DGS~\citep{kerbl20233d} and ContactGaussian-WM's spherical Gaussian representation, training all three methods using the same RGB-D inputs. Rendering quality is measured by averaging PSNR over 256 novel views per object. To evaluate geometric accuracy, we sample one million query points inside and outside each object and compute the RMSE between the Gaussian collision distances from Eq.~(\ref{eq:gs_sdf_single}) and the ground-truth mesh signed distances. As shown in Figure~\ref{fig:gaussian-quality}, our method substantially improves geometric accuracy over spherical 3DGS while maintaining high rendering quality, and outperforms ContactGaussian-WM on both metrics. These results demonstrate that our representation combines faithful appearance with accurate collision distances, supporting its use in subsequent visual tracking and physical simulation.

\vspace{-10pt}
\subsection{Q2: Trajectory Fidelity and Feasibility.}
\vspace{-5pt}
\subsubsection{Experimental Setup}
\begin{figure}[htbp]
\vspace{-5pt}
\centering
\includegraphics[width=\linewidth]{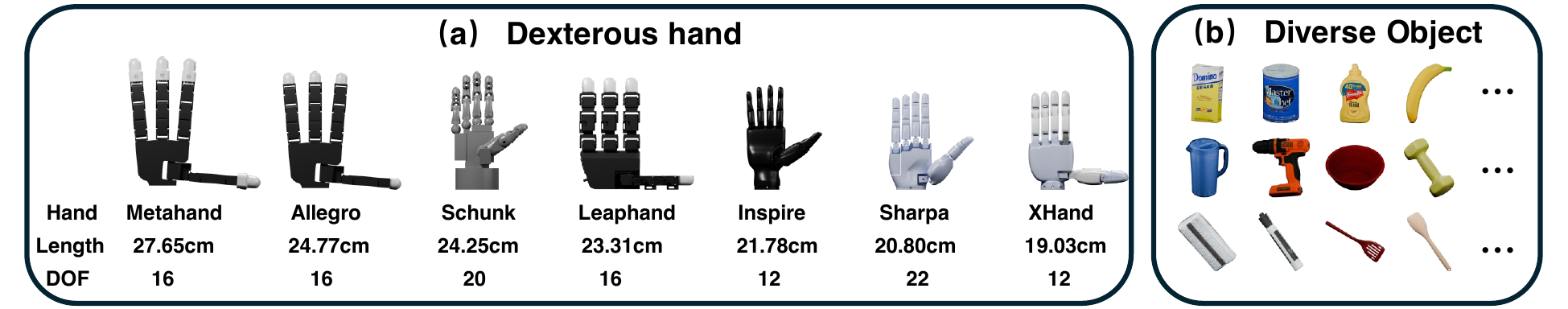}
\vspace{-15pt}
\caption{Dexterous hands and diverse objects used in evaluation. DexForge is evaluated on diverse objects and seven robotic hands that differ in size, degrees of freedom, and finger count.}
\label{fig:hand_object}
\vspace{-10pt}
\end{figure}

\vspace{-5pt}
\label{sec:Experimental Setup}
\noindent\textbf{Datasets and Robot Embodiments.} We evaluate dexterous retargeting on 130 sequences covering 40 objects: 50 right-hand grasping sequences from DexYCB~\citep{chao2021dexycb} and 80 precise in-hand manipulation sequences from HOT3D~\citep{banerjee2025hot3d}. Selection criteria are detailed in Appendix~\ref{app:dataset-selection}. We reconstruct spherical Gaussian models for all objects and recover object pose trajectories and corresponding MANO hand motion from the interaction videos provided by both datasets for subsequent retargeting. We use seven target hands---Allegro, LEAP, Sharpa, XHand, Inspire, MetaHand, and Schunk---that vary substantially in degrees of freedom, size, and finger configuration. Figure~\ref{fig:hand_object} shows the diverse objects and target robot hands.

\noindent\textbf{Evaluation Metrics.} Let $\boldsymbol{\rho}_t$ and $\mathbf{r}_t$ denote object position and unit quaternion; superscript $\mathrm{ref}$ denotes the dataset-provided or reconstructed reference. Mean tracking errors are $E_{\mathrm{pos}}=T^{-1}\sum_{t=1}^{T}\|\boldsymbol{\rho}_t-\boldsymbol{\rho}_t^{\mathrm{ref}}\|_2$ and $E_{\mathrm{rot}}=T^{-1}\sum_{t=1}^{T}\arccos(2\langle\mathbf{r}_t,\mathbf{r}_t^{\mathrm{ref}}\rangle^2-1)$. Contact error is $E_{\mathrm{contact}}^{\mathrm{refined}}=(\sum_{(i,\mathcal{I})}|\mathcal{I}|)^{-1}\sum_{(i,\mathcal{I})}\sum_{t\in\mathcal{I}}\|\mathbf{P}^{\mathrm{robot}}_i(q_t^{\mathrm{robot}})-q_t^{\mathrm{object}}\hat{\mathbf{a}}_{i,\mathcal{I}}\|_2$; $E_{\mathrm{contact}}^{\mathrm{raw}}$ instead uses original anchors $\mathbf{a}_{i,\mathcal{I}}$. A trajectory succeeds if $E_{\mathrm{rot}}<0.1\,\mathrm{rad}$ and $E_{\mathrm{pos}}<0.01\,\mathrm{m}$.

\vspace{-5pt}
\subsubsection{Ablation Study}
\vspace{-5pt}
To assess individual component contributions, we compare our full method with three cumulative ablation variants: (1) replacing reconstructed hand--object trajectories with dataset-provided ones; (2) further disabling contact refinement; and (3) finally removing raw contact guidance. We evaluate all variants on five DexYCB and five HOT3D sequences covering diverse objects and contact patterns, using seven robotic hands in ComFree-GS. Stochastic optimization is repeated with five random seeds, and Table~\ref{tab:retargeting-ablation} reports mean success rates. Detailed settings are provided in Appendix~\ref{app:ablation}.

\noindent\textbf{Ablation Analysis.} All variants retain dynamics optimization, as kinematic trajectories alone cannot ensure physical executability. With dataset ground-truth (GT) hand--object trajectories, raw contact guidance improves retargeting success, and contact refinement yields further gains. These improvements are consistent with the components' roles: guidance provides explicit targets for establishing and maintaining contacts, while refinement adapts these targets to the robot's morphology through joint anchor and configuration optimization, accounting for reachability and encouraging force closure. Replacing GT trajectories with our reconstructions largely preserves success rates, with small decreases for some hands, supporting the framework's effectiveness with reconstructed inputs.
\begin{table}[t]
\centering
\caption{Success rates (\%) across seven hands, averaged over five seeds. A: visual reconstruction ($\times$: dataset-provided GT hand--object trajectories); B: contact guidance; C: contact refinement. All variants retain both retargeting stages. Best results, including ties, are shaded and bolded.}
\label{tab:retargeting-ablation}
\small
\setlength{\tabcolsep}{2pt}
\renewcommand{\arraystretch}{0.8}
\begin{NiceTabular*}{\linewidth}{@{\extracolsep{\fill}}l*{10}{c}@{}}
\CodeBefore
% DexYCB：最优值及并列最优值
\rectanglecolor{gray!20}{5-5}{5-11}
\rectanglecolor{gray!20}{6-5}{6-6}
\rectanglecolor{gray!20}{6-8}{6-11}
% HOT3D：最优值及并列最优值
\rectanglecolor{gray!20}{9-5}{9-11}
\rectanglecolor{gray!20}{10-5}{10-6}
\rectanglecolor{gray!20}{10-9}{10-11}
\Body
\toprule
\multirow{2}{*}{Dataset}
& Reconstruction
& \multicolumn{2}{c}{Contact-aware}
& \multicolumn{7}{c}{Success rate (\%) $\uparrow$} \\
\cmidrule(lr){2-2}\cmidrule(lr){3-4}\cmidrule(l){5-11}
& A & B & C
& Allegro & Inspire & LEAP & MetaHand & Schunk & Sharpa & XHand \\
\midrule
\multirow{4}{*}{DexYCB}
& $\times$ (GT) & $\times$ & $\times$
& 44 & 36 & 24 & 32 & 40 & 44 & 48 \\
& $\times$ (GT) & $\checkmark$ & $\times$
& 64 & 56 & 52 & 60 & 72 & 64 & 76 \\
& $\times$ (GT) & $\checkmark$ & $\checkmark$
& \textbf{84} & \textbf{80} & \textbf{68} & \textbf{76}
& \textbf{80} & \textbf{80} & \textbf{84} \\
& $\checkmark$ & $\checkmark$ & $\checkmark$
& \textbf{84} & \textbf{80} & 64 & \textbf{76}
& \textbf{80} & \textbf{80} & \textbf{84} \\
\midrule
\multirow{4}{*}{HOT3D}
& $\times$ (GT) & $\times$ & $\times$
& 4 & 0 & 0 & 0 & 4 & 12 & 8 \\
& $\times$ (GT) & $\checkmark$ & $\times$
& 20 & 12 & 24 & 16 & 16 & 24 & 20 \\
& $\times$ (GT) & $\checkmark$ & $\checkmark$
& \textbf{32} & \textbf{16} & \textbf{32} & \textbf{28}
& \textbf{32} & \textbf{48} & \textbf{36} \\
& $\checkmark$ & $\checkmark$ & $\checkmark$
& \textbf{32} & \textbf{16} & 28 & 24
& \textbf{32} & \textbf{48} & \textbf{36} \\
\bottomrule
\end{NiceTabular*}
\end{table}

\vspace{-5pt}
\subsubsection{Retargeting Results}
\begin{figure}[htbp]
\vspace{-15pt}
\centering
\includegraphics[width=\linewidth]{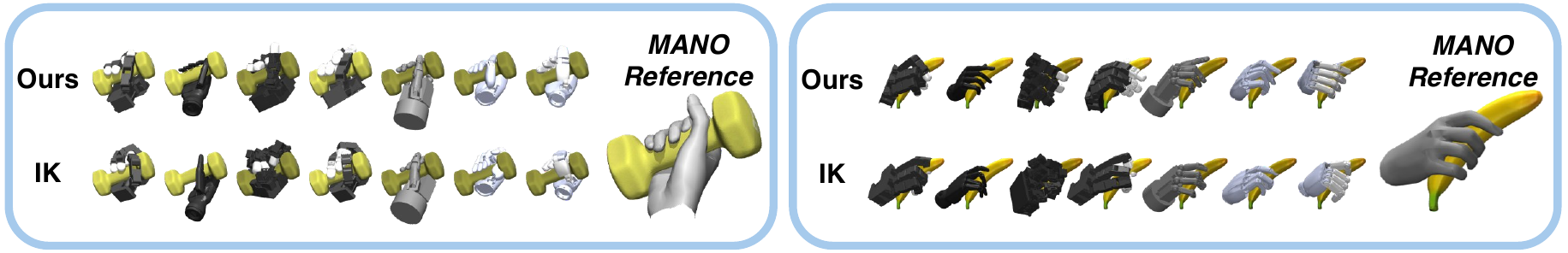}
\vspace{-15pt}
\caption{Visualization of selected kinematic retargeting results from Contact-aware and IK.}
\label{fig:kinematic}
\vspace{-15pt}
\end{figure}

\vspace{-5pt}
\label{lab:Contact-aware Kinematic Retargeting Results}
\noindent\textbf{Contact-aware Kinematic Retargeting Results.} We compare our method with SPIDER's IK-based stage~\citep{pan2026spiderscalablephysicsinformeddexterous} on all selected DexYCB and HOT3D sequences across seven hands. The baseline fits MANO-derived fingertip positions and wrist orientation without considering hand--object interactions or stable contacts. Both methods use dataset-provided GT trajectories for comparison; we additionally evaluate our method on our reconstructed trajectories. We report mean penetration $E_k=T^{-1}\sum_{t=1}^{T}\max(0,-\Phi_t^k)$, where $\Phi_t^k$ is the minimum hand--object ($k=\mathrm{ho}$) or self-collision ($k=\mathrm{self}$) distance at frame $t$ (Eq.~(\ref{eq:contact_qp})). Both methods are also evaluated using $E_{\mathrm{contact}}^{\mathrm{raw}}$ and $E_{\mathrm{contact}}^{\mathrm{refined}}$ from Section~\ref{sec:Experimental Setup}. Table~\ref{tab:kinematic-comparison} and Figure~\ref{fig:kinematic} show substantially reduced penetration and contacts better adapted to the robot while preserving the demonstrated contact structure. Comparable results with our reconstructions suggest limited impact on kinematic retargeting performance.
\begin{table}[t]
\centering
\caption{Kinematic retargeting errors (mm, mean $\pm$ SD). $E_{\mathrm{self}}$/$E_{\mathrm{ho}}$: self-collision/hand--object penetration; contact errors measure raw/refined anchor tracking. GT/recon.\ denote dataset-provided/our reconstructed trajectories. Best results with GT inputs are shaded and bolded.}
\label{tab:kinematic-comparison}
\small
\setlength{\tabcolsep}{2pt}
\renewcommand{\arraystretch}{0.8}
\begin{NiceTabular*}{\linewidth}{@{\extracolsep{\fill}}llcccc@{}}
\CodeBefore
\rectanglecolor{gray!20}{3-3}{3-6}
\rectanglecolor{gray!20}{6-3}{6-6}
\Body
\toprule
Dataset & Method
& $E_{\mathrm{self}}\downarrow$
& $E_{\mathrm{ho}}\downarrow$
& $E_{\mathrm{contact}}^{\mathrm{raw}}\downarrow$
& $E_{\mathrm{contact}}^{\mathrm{refined}}\downarrow$ \\
\midrule
\multirow{3}{*}{DexYCB}
& IK (GT)
& $4.73 \pm 5.02$
& $6.37 \pm 4.24$
& $23.30 \pm 32.58$
& $24.93 \pm 32.79$ \\
& Contact-aware (GT)
& $\boldsymbol{0.00 \pm 0.00}$
& $\boldsymbol{0.00 \pm 0.00}$
& $\boldsymbol{9.20 \pm 5.43}$
& $\boldsymbol{3.92 \pm 2.12}$ \\
\cmidrule(lr){2-6}
& Contact-aware (recon.)
& $0.00 \pm 0.00$
& $0.00 \pm 0.00$
& $10.22 \pm 7.64$
& $3.81 \pm 3.42$ \\
\midrule
\multirow{3}{*}{HOT3D}
& IK (GT)
& $5.37 \pm 5.30$
& $8.80 \pm 5.01$
& $23.72 \pm 13.16$
& $24.01 \pm 13.38$ \\
& Contact-aware (GT)
& $\boldsymbol{0.00 \pm 0.01}$
& $\boldsymbol{0.01 \pm 0.01}$
& $\boldsymbol{12.87 \pm 7.53}$
& $\boldsymbol{5.59 \pm 6.59}$ \\
\cmidrule(lr){2-6}
& Contact-aware (recon.)
& $0.01 \pm 0.01$
& $0.02 \pm 0.01$
& $15.24 \pm 8.91$
& $5.76 \pm 5.21$ \\
\bottomrule
\end{NiceTabular*}
\end{table}

\begin{figure}[htbp]
\centering
\vspace{-5pt}
\includegraphics[width=\linewidth]{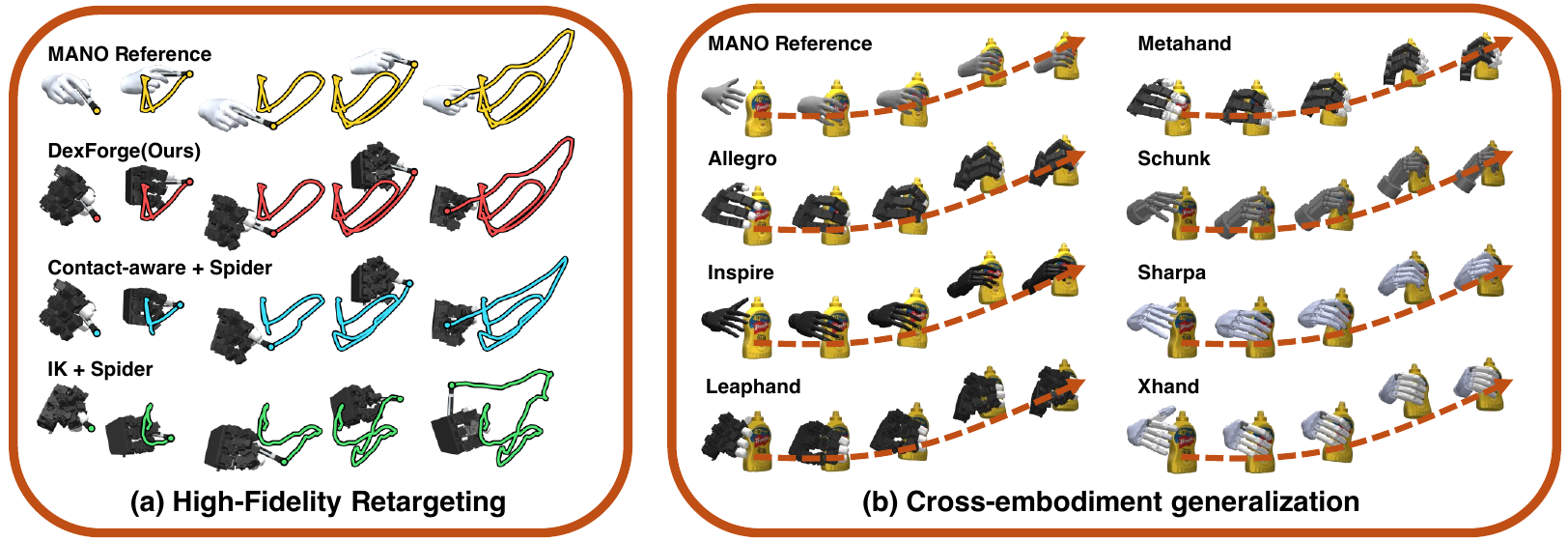}
\vspace{-15pt}
\caption{Qualitative results of force-aware dynamics retargeting. (a) Motion reproduction by our method and baselines relative to the MANO reference. (b) Retargeting across seven robotic hands.}
\label{fig:dynamics-retargeting}
\vspace{-5pt}
\end{figure}

\noindent\textbf{Force-Aware Dynamics Retargeting Results.} We compare DexForge with sampling-based SPIDER~\citep{pan2026spiderscalablephysicsinformeddexterous} on all selected sequences across seven hands in ComFree-GS. For a controlled comparison, Contact-aware+SPIDER replaces SPIDER's IK stage with our contact-aware kinematic stage and supplies our contact guidance to SPIDER's optimizer. Both baselines retain official SPIDER hyperparameters, and all methods allow a bounded initial hand-pose correction. All methods use dataset-provided GT trajectories; we additionally evaluate DexForge on our reconstructions. As shown in Table~\ref{tab:dynamics-retargeting} and Figure~\ref{fig:dynamics-retargeting}, Contact-aware+SPIDER improves over IK+SPIDER, particularly on DexYCB, showing that robot-adapted contact guidance also benefits sampling-based optimization. With the same kinematic stage and guidance, DexForge achieves the highest success rates and the lowest tracking errors on successful trajectories, a gain attributable to gradient-based force-aware refinement. The gap is largest on HOT3D, where objects are typically in hand from the first frame and SPIDER mostly fails through object drops, suggesting that simulation gradients help sustain precise in-hand contact. DexForge remains effective with reconstructed inputs, and visualizations show more faithful motion reproduction than the baselines across diverse hand morphologies. Hyperparameter settings, sensitivity analysis, and implementation details are in Appendix~\ref{app:Dynamics Retargeting Results}.
\begin{table}[t]
\centering
\caption{Dynamics retargeting results across seven hands (mean $\pm$ SD over five seeds). Position and rotation errors are computed over successful trajectories only. GT/recon.\ denote dataset-provided/our reconstructed trajectories. Best results with GT inputs are shaded and bolded.}
\label{tab:dynamics-retargeting}
\small
\setlength{\tabcolsep}{2pt}
\renewcommand{\arraystretch}{0.8}
\begin{NiceTabular*}{\linewidth}{@{\extracolsep{\fill}}llccc@{}}
\CodeBefore
\rectanglecolor{gray!20}{4-3}{4-5}
\rectanglecolor{gray!20}{8-3}{8-5}
\Body
\toprule
Dataset & Method
& Position error (mm) $\downarrow$
& Rotation error ($^\circ$) $\downarrow$
& Success rate (\%) $\uparrow$ \\
\midrule
\multirow{4}{*}{DexYCB}
& IK+SPIDER (GT)
& $8.21 \pm 1.22$
& $4.22 \pm 0.84$
& $24.9 \pm 1.6$ \\
& Contact-aware+SPIDER (GT)
& $7.57 \pm 1.43$
& $3.37 \pm 0.92$
& $50.6 \pm 2.1$ \\
& DexForge (GT)
& $\boldsymbol{2.73 \pm 1.78}$
& $\boldsymbol{0.93 \pm 0.70}$
& $\boldsymbol{77.4 \pm 1.6}$ \\
\cmidrule(lr){2-5}
& DexForge (recon.)
& $3.24 \pm 1.93$
& $0.97 \pm 0.63$
& $74.1 \pm 1.9$ \\
\midrule
\multirow{4}{*}{HOT3D}
& IK+SPIDER (GT)
& $7.55 \pm 1.74$
& $4.70 \pm 0.61$
& $0.4 \pm 0.1$ \\
& Contact-aware+SPIDER (GT)
& $7.07 \pm 2.06$
& $3.29 \pm 1.05$
& $0.6 \pm 0.2$ \\
& DexForge (GT)
& $\boldsymbol{5.01 \pm 2.55}$
& $\boldsymbol{1.34 \pm 0.95}$
& $\boldsymbol{34.9 \pm 2.1}$ \\
\cmidrule(lr){2-5}
& DexForge (recon.)
& $5.79 \pm 2.31$
& $1.68 \pm 0.97$
& $31.2 \pm 2.6$ \\
\bottomrule
\end{NiceTabular*}
\end{table}

\vspace{-10pt}
\subsection{Q3: Trajectory Transferability}
\vspace{-5pt}
\noindent\textbf{Sim-to-Sim Open-Loop Transfer.} We evaluate open-loop transfer from ComFree-GS to MuJoCo-CPU on the same subsets as the ablation study, with all methods using our reconstructed trajectories. The target uses mesh-based collision detection and object-specific physical parameters, while ComFree-GS uses Gaussian collisions and shared initial parameters. Per-sequence system identification calibrates the physical parameters and collision-distance offset $\delta$ to reduce dynamics mismatch and compensate for residual geometric error. We then reoptimize controls in the calibrated simulator and replay them open-loop in MuJoCo-CPU. We compare IK+SPIDER, Contact-aware+SPIDER, and our method, and evaluate TSMC against gradient-only and sampling-only identification. Transfer succeeds when the mean tracking errors satisfy $E_{\mathrm{pos}}<0.02\,\mathrm{m}$ and $E_{\mathrm{rot}}<0.2\,\mathrm{rad}$. Table~\ref{tab:sim-to-sim-transfer} shows that system identification improves transfer success across both datasets, with TSMC outperforming the two alternatives. Experimental details are provided in Appendix~\ref{app:sim-to-sim-transfer}.

\begin{table}[tbp]
\centering
\caption{Sim-to-sim transfer success rates (\%, mean $\pm$ SD across five seeds) before and after system identification. Best means are shaded and bolded.}
\label{tab:sim-to-sim-transfer}
\small
\setlength{\tabcolsep}{2pt}
\renewcommand{\arraystretch}{0.8}
\begin{NiceTabular*}{\linewidth}{@{\extracolsep{\fill}}llcccc@{}}
\CodeBefore
\rectanglecolor{gray!20}{5-3}{5-6}
\rectanglecolor{gray!20}{8-3}{8-6}
\Body
\toprule
\multirow{2}{*}{Dataset}
& \multirow{2}{*}{Retargeting method}
& \multirow{2}{*}{\shortstack{Before\\identification}}
& \multicolumn{3}{c}{After identification} \\
\cmidrule(lr){4-6}
& & & Gradient-only & Sampling-only & TSMC \\
\midrule
\multirow{3}{*}{DexYCB}
& IK+SPIDER
& $8.6 \pm 2.9$
& $11.4 \pm 2.9$
& $18.3 \pm 1.6$
& $22.3 \pm 3.7$ \\
& Contact-aware+SPIDER
& $15.4 \pm 1.6$
& $19.4 \pm 1.3$
& $31.4 \pm 2.0$
& $36.0 \pm 2.6$ \\
& DexForge
& $\boldsymbol{25.7 \pm 2.0}$
& $\boldsymbol{36.6 \pm 1.3}$
& $\boldsymbol{53.1 \pm 2.6}$
& $\boldsymbol{58.9 \pm 1.6}$ \\
\midrule
\multirow{3}{*}{HOT3D}
& IK+SPIDER
& $2.3 \pm 1.3$
& $4.6 \pm 1.6$
& $8.0 \pm 1.3$
& $14.3 \pm 2.9$ \\
& Contact-aware+SPIDER
& $4.0 \pm 1.6$
& $6.9 \pm 2.6$
& $12.0 \pm 2.4$
& $15.4 \pm 2.6$ \\
& DexForge
& $\boldsymbol{9.1 \pm 1.3}$
& $\boldsymbol{13.1 \pm 1.6}$
& $\boldsymbol{26.9 \pm 1.6}$
& $\boldsymbol{33.1 \pm 1.6}$ \\
\bottomrule
\end{NiceTabular*}
\end{table}

\begin{wrapfigure}{r}{0.5\textwidth}
\vspace{-10pt}
\centering
\includegraphics[width=\linewidth]{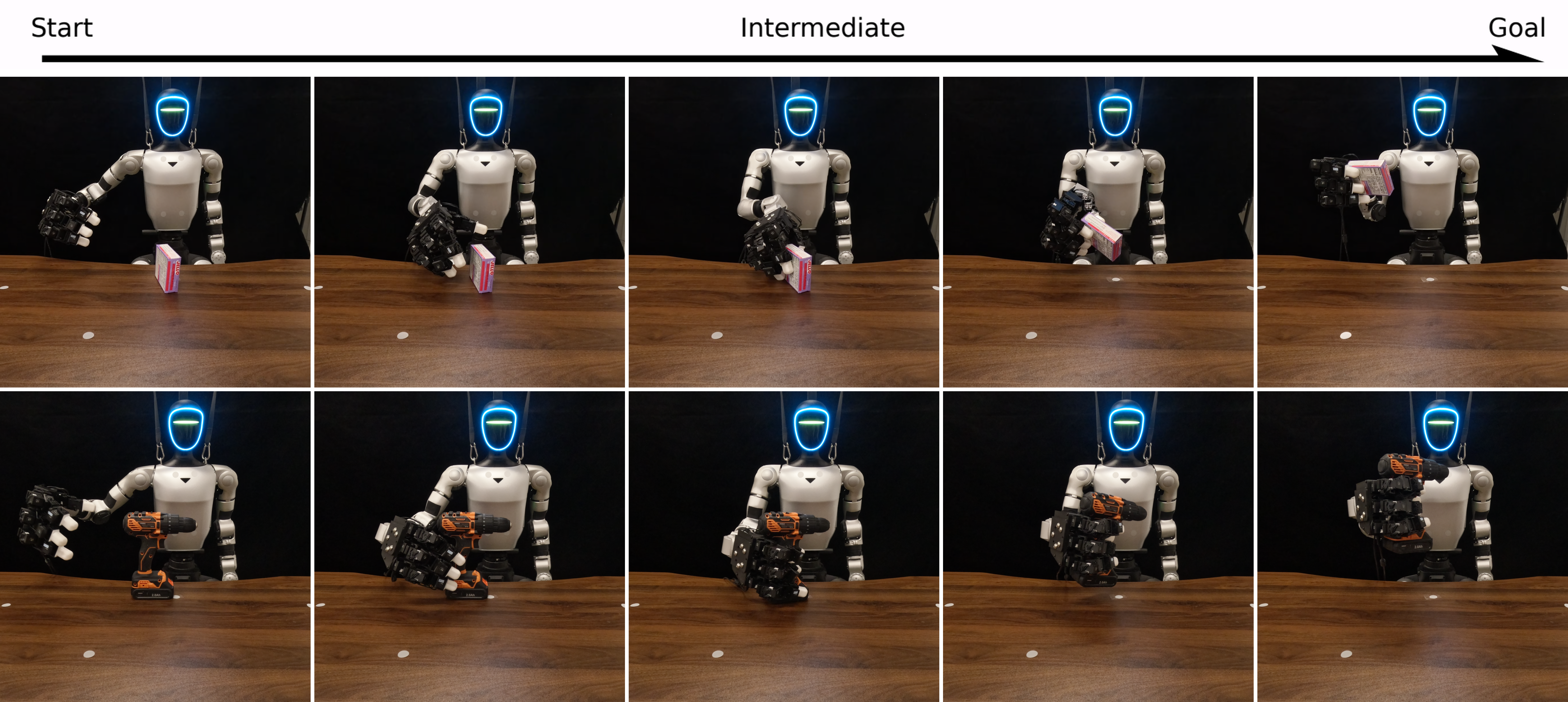}
\caption{Sim-to-real open-loop transfer. Controls optimized in ComFree-GS are directly replayed on a Unitree G1 equipped with a LEAP Hand v1.}
\label{fig:real_replay}
\vspace{-10pt}
\end{wrapfigure}
\noindent\textbf{Sim-to-Real Open-Loop Transfer.} Lacking accurate object-pose measurements, we evaluate transfer without system identification. For five YCB objects, we run the full retargeting pipeline on our reconstructions with known physical parameters. The resulting controls are replayed open-loop on a Unitree G1 with a LEAP Hand v1~\citep{shaw2023leap}, from simulation-matched initial positions. Over 10 trials per object, executions (Figure~\ref{fig:real_replay}) achieve an 80\% success rate in stable grasp-and-hold, which measures executability rather than tracking fidelity (Appendix~\ref{app:real-replay}).
\vspace{-10pt}
\section{Conclusion}
\vspace{-10pt}
DexForge connects visual reconstruction with Gaussian-based differentiable simulation to faithfully reproduce human manipulation from video demonstrations across diverse robot embodiments. Robot-adapted stable contacts guide both collision-constrained kinematic reference construction and gradient-based dynamics refinement, substantially improving retargeting success and object-tracking accuracy. Sim-to-sim transfer and real-world open-loop execution further demonstrate the applicability of the generated trajectories. These results highlight the potential of combining contact adaptation and differentiable physics to translate human manipulation experience into high-quality robot demonstrations for learning precise manipulation skills.

% \subsection*{AI Use Statement}
% Generative AI tools were used to improve the clarity and readability of the manuscript and to assist in retrieving and discovering related work. All retrieved references were verified by the authors against the original sources. The authors reviewed and edited all AI-assisted text and take full responsibility for the final content.

% \subsection*{Ethics Statement}
% This work adheres to the ICLR Code of Ethics. We use the publicly available DexYCB and HOT3D datasets in accordance with their respective licenses. We respect the privacy of individuals represented in the data and the conditions governing data use and redistribution. Our research aims to support the responsible development of dexterous robotic manipulation. The authors take full responsibility for the accuracy and integrity of the methods, results, and conclusions presented in this paper.

% \subsection*{Reproducibility Statement}
% The appendices provide dataset selection criteria, implementation details, optimization settings, and evaluation protocols. Our experiments use the publicly available DexYCB and HOT3D datasets. Additional results and a link to our anonymous code repository are available on our project website: \url{https://wmz1226.github.io/DexForge/}.

\clearpage
\bibliography{references}
\bibliographystyle{iclr2027_conference}

\appendix
\clearpage
\section{Gaussian Training Details}

\label{app:gaussian-training}
We jointly optimize Gaussian geometry and appearance using RGB reconstruction and signed-distance supervision from multi-view observations. This section describes reference SDF construction, supervision sampling, and the joint objective combining geometric and photometric losses.

\noindent\textbf{Reference SDF Construction.} We fuse multi-view depth maps into a truncated signed distance field (TSDF) and convert it into a reference Euclidean signed distance field $\phi^{\mathrm{ref}}$ following~\citep{Oleynikova_2017}. The reference field measures the shortest Euclidean distance to the reconstructed surface, with negative values inside the object and positive values outside, and provides geometric supervision for the Gaussian collision representation during training.

\noindent\textbf{Supervision Sampling.} We sample $10^6$ points near the object surface and query their reference distances $d_i=\phi^{\mathrm{ref}}(p_i)$. We store these pairs as $\mathcal{D}_{\mathrm{SDF}}=\{(p_i,d_i)\}_{i=1}^{10^6}$ and randomly draw mini-batches $\mathcal{B}\subset\mathcal{D}_{\mathrm{SDF}}$ during training, avoiding repeated reference-distance queries.

\noindent\textbf{Joint Training Objective.} We combine photometric and distance-matching losses:
\begin{equation}
\mathcal{L}=\mathcal{L}_{\mathrm{3DGS}}+\lambda_{\mathrm{SDF}}\mathcal{L}_{\mathrm{SDF}},\qquad \mathcal{L}_{\mathrm{SDF}}=\frac{1}{|\mathcal{B}|}\sum_{(p_i,d_i)\in\mathcal{B}}\left(\phi^{\mathrm{GS}}(p_i)-d_i\right)^2.
\label{eq:joint_gaussian_training}
\end{equation}
Here, $\mathcal{L}_{\mathrm{3DGS}}$ combines the standard $\ell_1$ and D-SSIM reconstruction losses, and $\lambda_{\mathrm{SDF}}$ balances geometric and photometric supervision. The distance loss updates Gaussian centers and scales, while the photometric objective updates both geometry and appearance.

\section{Hand and Object Motion Reconstruction}
\label{app:motion-reconstruction}

We reconstruct MANO hand motion and object poses from synchronized, calibrated multi-view RGB videos using the pretrained Gaussian object model $O$. Hand--object foreground masks are obtained with SAM~2~\citep{ravi2024sam2segmentimages} video segmentation, prompted once in the first frame of each view. This stage uses only RGB observations; depth supervision is used separately for object-model training.

\noindent\textbf{Multi-View MANO Fitting.} We detect right hands using the WiLoR~\citep{potamias2025wilorendtoend3dhand} hand detector and obtain per-view MANO estimates with HaMeR~\citep{pavlakos2023reconstructinghands3dtransformers}. Confidence-weighted rotation averaging and keypoint triangulation initialize the hand motion in world coordinates. We jointly optimize per-frame poses and translations with a shared hand shape, minimizing robust multi-view reprojection error, temporal acceleration, and deviations from the initial estimates. A strong pose prior limits articulation changes caused by noisy keypoint predictions. Optimization uses Adam for 800 iterations.

\noindent\textbf{Object Pose Tracking.} We extend the render-and-compare refinement of GS-Pose~\citep{cai2024gsposegeneralizablesegmentationbased6d} to multiple views while keeping Gaussian geometry and appearance fixed. The fitted MANO mesh provides hand masks that exclude occluded pixels. Object poses are optimized through differentiable rendering using silhouette overlap and photometric losses over the remaining pixels. The first frame uses multiple rotation hypotheses, while subsequent frames are initialized from the previous pose and a constant-velocity prediction. Coarse-to-fine refinement, rotation regularization, and loss-triggered re-initialization improve tracking robustness.

\noindent\textbf{Hand--Object Refinement.} With object poses fixed, we further refine the hand using the Gaussian collision distances in Eq.~\ref{eq:gs_sdf_pair}. Vertex--frame pairs within $5\,\mathrm{mm}$ of the object, including penetrating vertices, form a fixed candidate contact set. We augment the hand-fitting objective with squared penetration and contact-distance penalties, weighted by 10 and 1, respectively, with distances measured in millimeters, and optimize for another 400 iterations. The resulting MANO motion and object poses provide the reference trajectories for retargeting.

\begin{figure}[htbp]
\centering
\includegraphics[width=\linewidth]{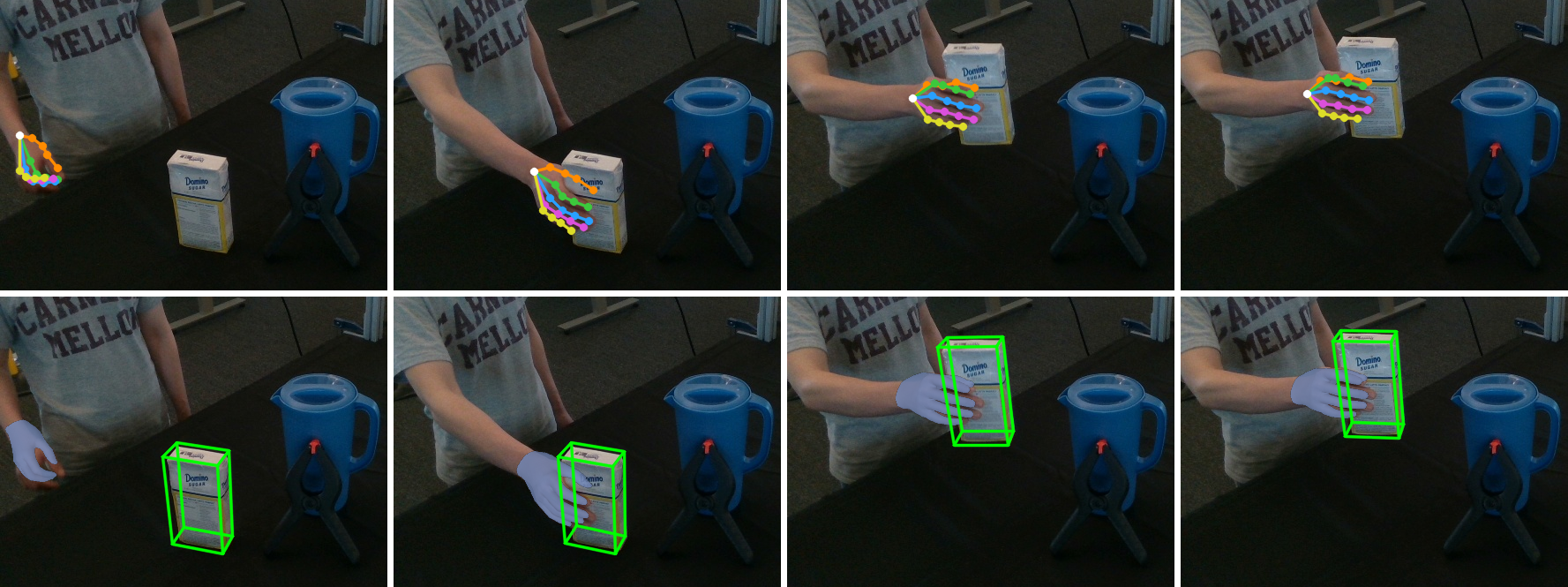}
\vspace{-15pt}
\caption{Visualization of the reconstructed hand motion and the estimated object pose on a DexYCB sequence, overlaid
on the original video from one camera view. Columns show four evenly spaced frames in temporal order. Top: the
reconstructed 21 hand keypoints projected into the image. Bottom: the reconstructed MANO hand mesh and the estimated
object pose, visualized as the green 3D bounding box of the object's Gaussian model.}
\label{fig:reconstruction}
\vspace{-10pt}
\end{figure}

\noindent\textbf{Reconstruction Evaluation.} We evaluate all 130 selected sequences using object translation and rotation errors, world-frame and wrist-aligned MPJPE, and hand--object penetration depth and penetrating-frame percentage. Table~\ref{tab:reconstruction} reports means and standard deviations across sequences, while Fig.~\ref{fig:reconstruction} shows a qualitative DexYCB example. Our method achieves accurate reconstruction with eight views on DexYCB and one RGB plus two grayscale SLAM views on HOT3D, with substantially less penetration than dataset annotations on both datasets.

\begin{table}[t]
\centering
\small
\setlength{\tabcolsep}{3pt}
\caption{Hand--object reconstruction results on real multi-view videos from DexYCB and HOT3D (mean $\pm$ SD across sequences). GT denotes dataset annotations. Lower is better for all metrics.}
\label{tab:reconstruction}
\resizebox{\linewidth}{!}{
\begin{tabular}{llcccccc}
\toprule
Dataset & Method & \multicolumn{2}{c}{Object pose} & \multicolumn{2}{c}{Hand accuracy} & \multicolumn{2}{c}{Geometric consistency} \\
\cmidrule(lr){3-4}\cmidrule(lr){5-6}\cmidrule(lr){7-8}
& & \shortstack{Translation\\(mm)}
& \shortstack{Rotation\\($^\circ$)}
& \shortstack{World-frame\\MPJPE (mm)}
& \shortstack{Wrist-aligned\\MPJPE (mm)}
& \shortstack{Penetration\\depth (mm)}
& \shortstack{Penetrating\\frames (\%)} \\
\midrule
\multirow{2}{*}{DexYCB}
& GT   & $0.00 \pm 0.00$ & $0.00 \pm 0.00$ & $0.00 \pm 0.00$ & $0.00 \pm 0.00$ & $3.15 \pm 1.79$ & $59.5 \pm 13.6$ \\
& Ours & $0.40 \pm 0.16$ & $0.46 \pm 0.38$ & $6.42 \pm 1.18$ & $4.82 \pm 0.86$ & $0.24 \pm 0.15$ & $3.9 \pm 8.6$ \\
\midrule
\multirow{2}{*}{HOT3D}
& GT   & $0.00 \pm 0.00$ & $0.00 \pm 0.00$ & $0.00 \pm 0.00$ & $0.00 \pm 0.00$ & $6.13 \pm 3.04$ & $97.8 \pm 7.3$ \\
& Ours & $12.23 \pm 3.11$ & $4.83 \pm 0.78$ & $23.47 \pm 2.34$ & $9.21 \pm 1.21$ & $1.38 \pm 0.21$ & $7.8 \pm 4.7$ \\
\bottomrule
\end{tabular}
}
\end{table}

\section{Gaussian Collision Detection}

\label{app:bvh}

We describe the spatial acceleration, contact geometry, differentiation, and simulator integration of Gaussian-based collision detection. Each object is represented by spherical Gaussian primitives rigidly attached to its body, with collision
radii $r_i=3\sigma_i$. Robot surface query points are represented as zero-radius primitives and handled by the same collision formulation. All centers used in the following distance and contact formulas are expressed in a common world coordinate
frame. We first describe the default hard contact mode, which is used for all results in this paper, and then an optional soft mode.

\noindent\textbf{BVH Construction.} We construct binary bounding volume hierarchies (BVHs)~\citep{bergen1997efficient} over each object's primitives in its body-local coordinate frame using a binned surface-area heuristic. Each node stores an
axis-aligned bounding box of the primitive centers and their maximum collision radius. The distance between two center bounding boxes, minus their maximum radii, provides a conservative lower bound on the corresponding sphere-pair distances. The
hierarchies are constructed during model loading and reused under rigid-body motion. During traversal, relative body transforms are applied to express the bounding boxes in a common coordinate frame.

\noindent\textbf{Distance Evaluation.} Collision queries are performed for each source--target primitive-group pair; an ungrouped object is treated as a single group. Let $\mathcal{I}_{\mathcal{A}}$ and $\mathcal{I}_{\mathcal{B}}$ denote the
primitive index sets of the queried groups. We compute the sphere-pair separation and select the minimum-distance pair:
\begin{equation}
\begin{aligned}
d_{ij}
&=
\left\|\boldsymbol{\mu}^{\mathcal{B}}_j-\boldsymbol{\mu}^{\mathcal{A}}_i\right\|_2
-r^{\mathcal{A}}_i-r^{\mathcal{B}}_j,\\
(i^\star,j^\star)
&=
\operatorname*{arg\,min}_{i,j} d_{ij},
\qquad
i\in\mathcal{I}_{\mathcal{A}},\quad
j\in\mathcal{I}_{\mathcal{B}}.
\end{aligned}
\label{eq:selected_sphere_pair}
\end{equation}
Brute-force evaluation of Eq.~(\ref{eq:gs_sdf_pair}) requires $O(N^{\mathcal{A}}N^{\mathcal{B}})$ pairwise distance computations. Joint BVH traversal maintains the best sphere-pair distance found so far and prunes node pairs whose conservative
bounds cannot improve it, subject to a numerical tolerance. Unpruned leaf-node pairs are evaluated by enumerating their sphere pairs. The initial search cutoff is determined by the contact-generation threshold. This procedure accelerates the
minimum-distance search over the queried groups without materializing all primitive pairs. Each group pair contributes at most one contact, and no contact is generated when no sphere pair satisfies the generation threshold.

\noindent\textbf{Contact Points and Normals.} For the selected sphere pair, we define the relative center displacement and a regularized direction from $\mathcal{A}$ to $\mathcal{B}$:
\begin{equation}
\mathbf{v}
=
\boldsymbol{\mu}^{\mathcal{B}}_{j^\star}
-
\boldsymbol{\mu}^{\mathcal{A}}_{i^\star},
\qquad
\mathbf{n}
=
\frac{\mathbf{v}}
{\sqrt{\mathbf{v}^{\top}\mathbf{v}+\epsilon^2}},
\quad \epsilon=10^{-9}.
\label{eq:pair_contact_geometry}
\end{equation}
The contact distance and representative contact point are computed as
\begin{equation}
\phi
=
d_{i^\star j^\star}+\Delta,
\qquad
\mathbf{p}
=
\boldsymbol{\mu}^{\mathcal{B}}_{j^\star}
-r^{\mathcal{B}}_{j^\star}\mathbf{n}
+\Delta\mathbf{n},
\label{eq:selected_contact_geometry}
\end{equation}
where $\Delta$ is a distance offset, set to zero by default. With $\Delta=0$ and noncoincident centers, $\mathbf{p}$ approximates the point on the selected target sphere facing the source sphere. This point is not guaranteed to lie on the outer
boundary of the complete sphere union. The contact frame is constructed from $\mathbf{n}$, with its orientation adjusted to match the geometry ordering expected by the dynamics solver. The regularizer prevents division by zero in the direction
computation, although coincident centers remain a degenerate contact configuration.

\noindent\textbf{Differentiation.} In the differentiable simulation path, BVH traversal and minimum-distance pair selection are performed outside the differentiation graph. The selected primitive indices are then held fixed while their
world-space centers, contact distance, contact point, and contact frame are recomputed within the differentiable computation. Gradients propagate through these continuous quantities to body poses and, through the dynamics, to the optimized
controls. We do not differentiate through BVH traversal, discrete pair selection, or contact activation decisions. For a unique minimizing pair and nondegenerate contact geometry, this computation differentiates the selected contact branch; pair
switches and contact activation changes can introduce nonsmooth transitions.

\noindent\textbf{Optional Soft Contact.} The formulation above selects a single sphere pair, so the contact point and normal change discontinuously when the minimizing pair switches. We additionally provide an optional soft mode that fuses all
nearby pairs of a group pair into one contact. Let $\mathcal{S}$ index the sphere pairs of the queried groups, with separations $d_k$, pairwise contact points $\mathbf{p}_k$, and normals $\mathbf{n}_k$ computed as in
Eqs.~(\ref{eq:pair_contact_geometry})--(\ref{eq:selected_contact_geometry}). The fused distance is the smooth minimum
\begin{equation}
\phi_{\mathrm{soft}}
=
\min_{\mathbf{w}\ge 0,\ \sum_k w_k=1}
\ \sum_{k\in\mathcal{S}} w_k d_k
+\frac{2h}{3}\Big(\sum_{k\in\mathcal{S}} w_k^{3/2}-1\Big),
\label{eq:soft_distance}
\end{equation}
with width $h=1$\,mm. Its minimizer has the closed form
\begin{equation}
w_k=a_k^2,
\qquad
a_k=\max\!\Big(\frac{\tau-d_k}{h},\,0\Big),
\qquad
\sum_{k\in\mathcal{S}} a_k^2=1,
\label{eq:soft_weights}
\end{equation}
where the threshold $\tau$ is the unique root of the last condition, which gives $\phi_{\mathrm{soft}}=\tau-\tfrac{h}{3}\big(\sum_k a_k^3+2\big)$. Only pairs with $d_k<\tau$ receive nonzero weight, and $\tau\le \min_k d_k+h$, so the support is
confined to pairs within $h$ of the closest one; with a single supporting pair the hard contact is recovered exactly, and in general $\min_k d_k-\tfrac{2h}{3}\le\phi_{\mathrm{soft}}\le\min_k d_k$. The contact point and normal are the weighted
combinations
\begin{equation}
\mathbf{p}_{\mathrm{soft}}=\sum_{k\in\mathcal{S}} w_k\,\mathbf{p}_k,
\qquad
\mathbf{n}_{\mathrm{soft}}=
\frac{\sum_{k\in\mathcal{S}} w_k\,\mathbf{n}_k}
{\big\|\sum_{k\in\mathcal{S}} w_k\,\mathbf{n}_k\big\|_2}.
\label{eq:soft_geometry}
\end{equation}
The support is collected with the same BVH: the nearest pairs seed the search, and a range query bounded by $\tau$ adds the remaining supporting pairs. Since $\partial\phi_{\mathrm{soft}}/\partial d_k=w_k$ and pairs enter and leave the support
with zero weight, the fused distance is continuously differentiable in the pair separations, and gradients are propagated through the distance, contact point, and normal. Each group pair still contributes at most one contact. The soft mode is
provided as an option in the released code for settings in which the hard contact gradients become unreliable, for example when the minimizing pair switches frequently along a rollout: every switch changes the contact point and normal
discontinuously, and over long differentiation horizons these nonsmooth events accumulate. The soft mode removes this particular source of nonsmoothness; it does not address the gradient growth or decay inherent to long horizons, and we have not
evaluated it on our benchmark.

\noindent\textbf{Simulator Integration.} We provide JAX and Warp collision backends for MJX and MJWarp, respectively. Our scene-loader extensions accept Gaussian collision declarations such as \verb|<geom type="gs" file="model.ply"/>| and load
the corresponding primitive parameters. The collision routine supplies distances, contact points, and oriented contact frames through MuJoCo-style contact interfaces~\citep{todorov2012mujoco}, together with the contact parameters specified by
the scene. These extensions modify geometry loading and collision dispatch while preserving the remaining scene configuration, downstream solver, and simulation loop. We integrate the routine into ComFree-Sim~\citep{borse2026comfree} to obtain
ComFree-GS.

\begin{wrapfigure}{r}{0.48\textwidth}
\vspace{-15pt}
\centering
\includegraphics[width=\linewidth]{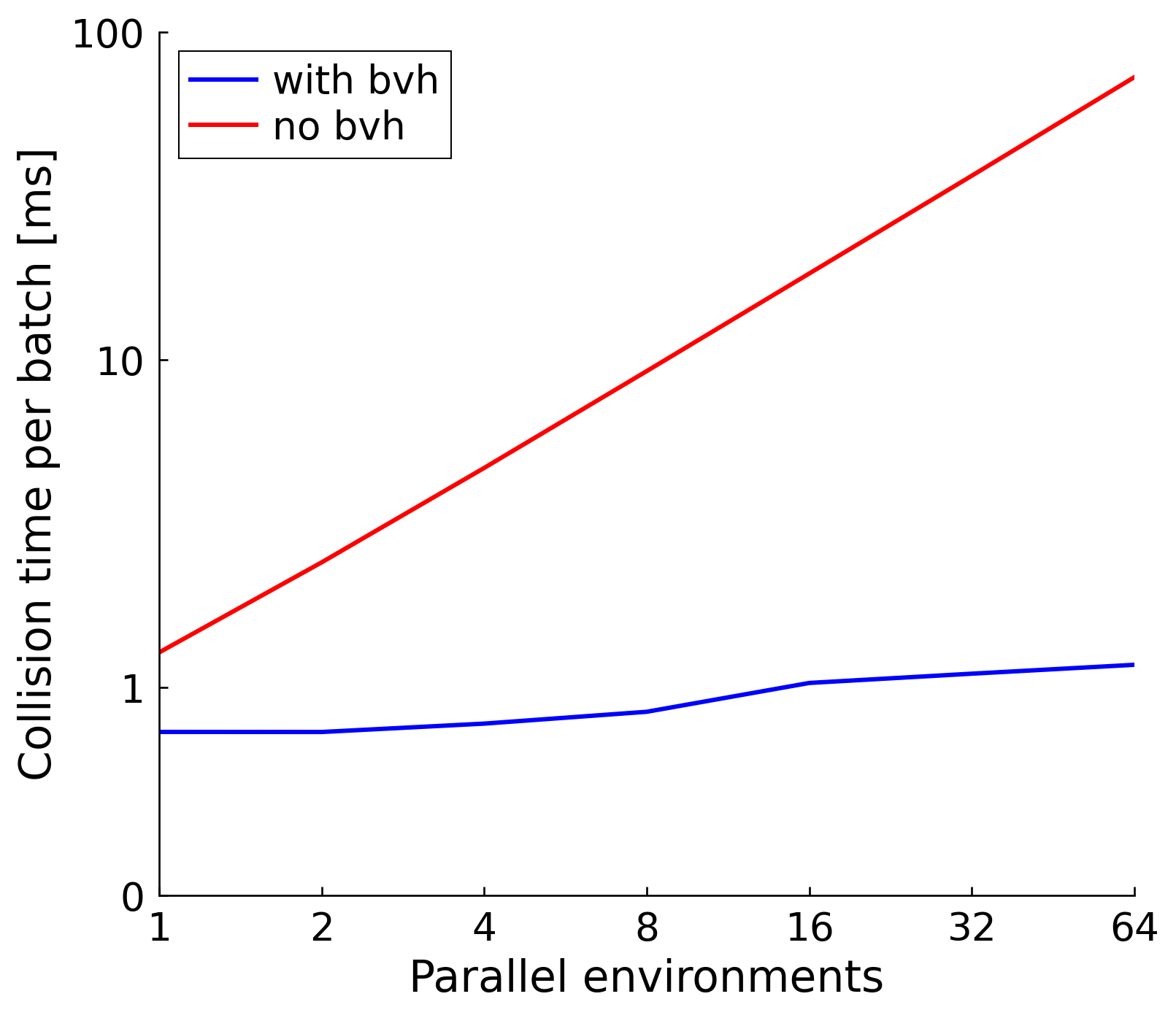}
\setlength{\abovecaptionskip}{-10pt}
\caption{Forward collision-query time with and without BVH for 1--64 parallel environments on an NVIDIA RTX 4090D GPU.}
\label{fig:collision_time_vs_worlds}
\vspace{-18pt}
\end{wrapfigure}
\noindent\textbf{Effect of BVH Acceleration.} We compare BVH-based queries with GPU brute-force evaluation in a fixed hand--object scene on an NVIDIA RTX 4090D GPU. Each environment contains 2,000 query points sampled from the robot hand surface
and an object represented by 10,000 spherical Gaussians. Both methods use identical geometry and poses to ensure a controlled comparison. Figure~\ref{fig:collision_time_vs_worlds} reports the forward collision-query time per batch for 1--64
parallel environments. Brute-force runtime grows approximately linearly with the number of environments, whereas BVH runtime increases only modestly across the tested batch sizes. BVH achieves speedups of $1.75\times$ and $62.09\times$ at 1 and
64 environments, respectively, while producing identical contact outputs. Timings exclude backward computation, BVH construction, compilation, and dynamics solving.

\section{Stable Contact Extraction and Robot Adaptation}
\label{app:contact-refinement}

\subsection{Stable Contact Extraction}
For each finger, the candidate contact points are the MANO vertices on the palmar side of its distal pad. At each frame $t$, we query their Gaussian distances at the reference object pose $q_t^{\mathrm{ref,object}}$ and project them onto the Gaussian surface in the object frame. A vertex is in contact when its distance lies within $\pm10\,\mathrm{mm}$, and contact runs shorter than eight frames at $30\,\mathrm{Hz}$ are discarded. Each run is split into maximal intervals $\mathcal{I}$ over which the projected points stay consistent, meaning that at least $80\%$ of them lie within $8\,\mathrm{mm}$ of their medoid; this medoid defines the anchor $\mathbf{a}_{j,\mathcal{I}}$. For each finger, we then select a non-overlapping set of these candidates that covers the most frames, penalizing switches between candidates. Each selected candidate yields a triplet $(j,\mathbf{a}_{j,\mathcal{I}},\mathcal{I})$, so a finger can have multiple anchors at different stages of the demonstration.

\subsection{Joint Contact and Configuration Optimization}
\noindent\textbf{Keyframes and Initialization.} We split the trajectory whenever a stable contact starts or ends and select one keyframe per segment: frame $0$ for the initial segment and the middle frame otherwise. We also add frames $t-1$, $t$, and $t+1$ around each transition at $t$ to support smooth configuration changes; the deduplicated frames form $\mathcal{K}$. All keyframes are initialized from a single seed, computed at the first frame with the most fingers in stable contact. This seed fits the robot fingertip pads to the anchors and their normals under joint limits and collision constraints, starting from several postures aligned with the MANO palm, and is repaired to the nearest collision-free configuration when needed. DexYCB keyframes reuse the seed directly, whereas for HOT3D the seed palm follows each keyframe's MANO palm.

\noindent\textbf{Anchor Regularization.} The anchor loss $\mathcal{L}_{\mathrm{a}}$ aggregates contributions over keyframes with active contacts. For a keyframe with $n$ active anchors, its contribution is
\[
\frac{\|\bar{\hat{\mathbf{a}}}-\bar{\mathbf{a}}\|_2^2}{R^2}
+\frac{1}{nR^2}\sum_i
\|(\hat{\mathbf{a}}_i-\bar{\hat{\mathbf{a}}})-(\mathbf{a}_i-\bar{\mathbf{a}})\|_2^2,
\]
where bars denote centroids and $R$ is the object radius. The two terms penalize centroid displacement and deviations from the original relative anchor layout, respectively, both with unit weight; the latter is averaged over active anchors.

\noindent\textbf{Force-Closure Residuals.} We distinguish incomplete contact sets, which contain fewer than three contacts or are strict subsets of larger observed contact sets, from complete sets, which are maximal contact sets with at least three contacts. We use $\mu=0.2$ throughout. For an incomplete set, each contact force is represented by four linearized friction-cone edges:
\[
\mathbf{f}_i=\sum_{\ell=1}^{4}\beta_{i\ell}\mathbf{e}_{i\ell},
\qquad
\mathbf{e}_{i\ell}=\frac{\mathbf{n}_i+\mu\mathbf{t}_{i\ell}}{4},
\qquad
1\leq\beta_{i\ell}\leq21,
\]
where $\mathbf{n}_i$ is the contact normal and $\mathbf{t}_{i\ell}$ are tangential directions. The coefficient bounds correspond to $F_{\max}=20$ plus a lower bound of $1$, which excludes the trivial zero-force solution. We solve bounded least squares to minimize the six-dimensional net-wrench residual
\[
\mathbf{w}=
\begin{bmatrix}
\sum_i\mathbf{f}_i\\
\sum_i\big((\mathbf{p}_i-\mathbf{c}_{\mathrm{obj}})/R\big)\times\mathbf{f}_i
\end{bmatrix},
\]
where $\mathbf{p}_i$ is the contact position and $\mathbf{c}_{\mathrm{obj}}$ is the object center of mass. The moment arms are normalized by $R$.

For a complete contact set, we solve a second-order cone program:
\[
\max_{\{\mathbf{f}_i\},\,r}\quad r
\qquad
\mathrm{s.t.}\quad
\mathbf{w}=0,\qquad
\sum_i f_{i,\mathrm{n}}=1,\qquad
\mu f_{i,\mathrm{n}}-\|f_{i,\mathrm{t}}\|_2\geq r
\quad\forall i,
\]
where $f_{i,\mathrm{n}}$ and $f_{i,\mathrm{t}}$ are the normal and tangential force components. Normalizing the total normal force fixes the force scale. The scalar residual is $\exp(-r/(2\mu))$, with a linear continuation when $r/(2\mu)<-3$. Both residual types enter the least-squares objective with weights proportional to the duration of the corresponding contact state; the shared force-closure weight is $1$.   Both residuals are differentiated analytically with respect to the contact points and normals, and chained through the forward-kinematics Jacobians of the robot pad points and the derivative of the nearest-sphere Gaussian normal; the anchors do not enter
$\mathcal{L}_{\mathrm{fc}}$. For the bounded least-squares residual, bound-active coefficients are held fixed and the free coefficients are differentiated implicitly through their normal equations. For the margin $r$, we use the envelope derivative given by
the optimal primal and dual solutions of the SOCP, which accounts for the redistribution of the optimal forces.

\noindent\textbf{Regularization.} We use $\mathcal{L}_{\mathrm{reg}}$ for the regularization terms in both optimization stages, with stage-specific targets and active terms. Per-frame residuals use trapezoidal temporal weights $\Delta t$ over the relevant keyframes or trajectory frames. Positions and rotations are expressed in meters and radians, respectively.

The MANO palm frame is centered at the mean position of the wrist and tracked finger-base joints. Its $x$-axis points from the little-finger base to the index-finger base, using the ring-finger base when the little finger is absent; its $z$-axis is perpendicular to the $x$-axis and the wrist-to-mean-fingertip direction. Palm-position tracking has weight $10$, while the rotation residual $\log((\mathbf{R}^{*})^\top\mathbf{R})$ has weight $1$, where $\mathbf{R}^{*}$ is the target orientation. At keyframe $k$, the target is the MANO palm pose and both weights are multiplied by $1-0.99c_k$, where $c_k=\max_i\alpha_{t_k,i}$.

Hand-shape residuals compare unit phalanx directions expressed in the respective robot and MANO palm frames. Each finger has weight $0.5(1-\alpha_f)$, where $\alpha_f$ is its selected contact's blending weight and is zero for fingers without stable contacts. The contact-normal residual is $r_i(\mathbf{n}_{\mathrm{robot}}+\mathbf{n}_{\mathrm{object}})$ with weight $100\alpha_{t,i}$, encouraging opposing contact normals. Here, $r_i$ is the RMS radius of the fingertip surface, converting the normal discrepancy to a positional scale.

For temporal regularization, let $\mathbf{z}$ stack the palm position, the palm rotation vector relative to the seed orientation, and the finger joint angles. We penalize
\[
\tau^2\int\|\dot{\mathbf{z}}\|_{\mathbf{W}}^2\,dt
+\tau^4\int\|\ddot{\mathbf{z}}\|_{\mathbf{W}\mathbf{M}}^2\,dt,
\qquad \tau=0.1\,\mathrm{s},
\]
using nonuniform finite differences. The diagonal entries of $\mathbf{W}$ are $100$ for palm translation and $1$ for palm rotation and finger joints. The corresponding entries of $\mathbf{M}$ are $32$ for palm coordinates and $16$ for finger joints. The same temporal operator is used over keyframes and over the full trajectory. Keyframe optimization additionally includes the proximity prior $10^{-3}\|(q-q_{\mathrm{init}})/s\|_2^2$, where division is componentwise and the translation components are normalized by the object radius.

\noindent\textbf{Solver.} We solve Eq.~\ref{eq:contact-refinement} using Gauss--Newton SQP within a trust region. Violated hand--object constraints are added as they appear, elastic slack handles inconsistent linearizations, and a backtracking line search with second-order corrections controls the updates. Anchors are retracted onto the Gaussian surface at every trial, and geometric feasibility is checked with a tolerance of $0.1\,\mathrm{mm}$. We return the best feasible iterate, or the least-violating one otherwise.

\subsection{Kinematic Trajectory Optimization}
\noindent\textbf{Contact Targets.} For each finger and frame, we select the temporally nearest stable-contact interval. The distance is zero inside an interval and otherwise equals the distance to its nearest endpoint; ties are resolved in favor of the
earlier interval. Both the anchor and robot contact query point are taken from the selected interval. The blending weight $\alpha_{t,i}$ is one within that interval and zero elsewhere. The contact-position fitting weight is $100(1+99\alpha_{t,i})$. Fingers
without stable-contact intervals instead track the MANO fingertips with weight $1$ in the position-fitting term.

\noindent\textbf{Full-Trajectory Regularization.} In Eq.~\ref{eq:contact_qp}, $\mathcal{L}_{\mathrm{reg}}$ applies the palm, hand-shape, and contact-normal penalties at every frame, together with the temporal operator over the full trajectory. The palm target is the MANO palm pose composed with spline-interpolated keyframe palm offsets. The keyframe proximity prior is not used in this stage.

\noindent\textbf{Initialization and Solver.} The trajectory is initialized by interpolating the keyframes; frames without demonstrated contact, such as approach and release, instead follow the MANO motion while matching every keyframe exactly. Keyframes
that violate the collision constraints are released, and the remaining keyframes, indexed by $\mathcal{K}_{\mathrm{f}}$, are fixed. After repairing the initial trajectory to be collision-free at every frame, we run up to $40$ Gauss--Newton SQP iterations
with a trust region and an Armijo line search. Collision checks are performed at every frame and, where applicable, at $16$ substeps between consecutive frames. Substep checks are skipped for intervals with residual between-frame collisions after repair, which are reported, and for intervals between adjacent fixed keyframes. Trial updates are accepted only if they pass all frame-wise checks and all applicable substep checks, maintaining frame-wise collision feasibility. For DexYCB, the hand and object are shifted vertically together so that the object rests on the table, and the object is then settled once in simulation. The rigid transformation from the pre-settling to the settled object pose is then applied uniformly to the entire hand--object trajectory.
\section{Force-Aware Optimization Details}
\label{app:forceaware}

We optimize position-control knots $\mathbf{U}=\mathbf{U}^{\mathrm{ref}}+\Delta\mathbf{U}$, which are linearly interpolated over the $L$ state-transition steps in Eq.~(\ref{eq:mpc}). The tracking residual is $\mathbf{e}_m=\operatorname{col}(q_m^{\mathrm{robot}}-q_m^{\mathrm{robot,ref}},\mathbf{p}_m^{\mathrm{object}}-\mathbf{p}_m^{\mathrm{object,ref}},\vartheta_m)$, where $\vartheta_m$ is the relative object rotation angle. Let $\eta_t$ equal one in the first optimization window and zero otherwise. Subject to the simulated dynamics, the complete objective is
\begin{equation}
\label{eq:forceaware_full}
\begin{aligned}
\min_{\mathbf{U},\mathbf{z}}\quad J_t ={} & \|\mathbf{e}_{t+L}\|_{\mathbf{Q}_{\mathrm{f}}}^{2}+\sum_{m=t+1}^{t+L-1}\|\mathbf{e}_m\|_{\mathbf{Q}}^{2} \\
&+\sum_{m=t+1}^{t+L}\left(\ell_{\mathrm{a},m}+\ell_{\mathrm{d},m}\right)+\sum_{m=t}^{t+L}\ell_{\mathrm{col},m} \\
&+\mathcal{R}(\mathbf{U},\mathbf{x})+\eta_t\lambda_{\mathrm{init}}\|\mathbf{b}\odot\tanh(\mathbf{z})\|_2^2.
\end{aligned}
\end{equation}
The initial hand configuration is $q_t^{\mathrm{robot}}=\bar q_t^{\mathrm{robot}}+\eta_t\mathbf{b}\odot\tanh(\mathbf{z})$, where $\bar q_t^{\mathrm{robot}}$ is the supplied hand configuration and $\mathbf{b}$ specifies componentwise correction bounds. The variable $\mathbf{z}$ is optimized only in the first window; the remaining initial-state components are fixed. Subsequent windows start from the current executed state. Terminal tracking replaces the final intermediate tracking term. All weights absorb the corresponding unit conversions and component normalization.

For correspondence $k=(i,\mathcal{I})$, define $\mathbf{r}_{k,m}=\mathbf{P}^{\mathrm{robot}}_k(q_m^{\mathrm{robot}})-\mathbf{T}^{\mathrm{object}}_m\hat{\mathbf{a}}_k$, where the robot material point is evaluated by forward kinematics and the refined anchor is transformed by the simulated object pose. The contact and collision terms are
\begin{equation}
\label{eq:forceaware_contact}
\begin{aligned}
\ell_{\mathrm{a},m} &= \lambda_{\mathrm{a}}\sum_k w^{\mathrm{a}}_{k,m}\gamma_{k,m}\|\mathbf{r}_{k,m}\|_2^2, \\
\ell_{\mathrm{d},m} &= \lambda_{\mathrm{d}}\sum_k w^{\mathrm{d}}_{k,m}[\phi_{k,m}-\phi_{\star}]_+^2, \\
\ell_{\mathrm{col},m} &= \sum_b\left(\lambda_{\mathrm{s}}[-\phi_{b,m}-d_{\mathrm{s}}]_+^2+\lambda_{\mathrm{p}}[-\phi_{b,m}-d_{\mathrm{p}}]_+^2\right) \\
&\quad+\lambda_{\mathrm{f}}\sum_{(a,b)\in\mathcal{P}}[\epsilon_{\mathrm{f}}-d_{ab,m}]_+^2.
\end{aligned}
\end{equation}
Here, $[z]_+=\max(z,0)$, and $w^{\mathrm{a}}_{k,m}$ and $w^{\mathrm{d}}_{k,m}$ incorporate guidance activation, confidence, and contact-onset ramps. The distance-dependent anchor weight is $\gamma_{k,m}=\sigma((\|\mathbf{r}_{k,m}\|_2-d_0)/\tau)$, where $\sigma$ is the sigmoid function; it is recomputed during forward evaluation and held fixed during the corresponding backward pass. The quantity $\phi_{k,m}$ is the simulator collision distance for the hand collision output associated with correspondence $k$, rather than the material-point anchor distance, and $\phi_{\star}<0$ specifies the target contact compression. The penetration term covers hand--object collision outputs, with thresholds $d_{\mathrm{s}}=-\phi_{\star}$ and $d_{\mathrm{p}}>d_{\mathrm{s}}$. The set $\mathcal{P}$ contains selected inter-finger capsule pairs, $d_{ab,m}$ denotes their surface separation, and $\epsilon_{\mathrm{f}}$ is the desired margin. These are soft penalties. Their weights follow the selected control profile and may be zero when a term is disabled.

The smoothness regularizer penalizes control-knot differences and simulated hand velocities and accelerations:
\begin{equation}
\label{eq:forceaware_smooth}
\begin{aligned}
\mathcal{R} &= \sum_{k=1}^{K}\left(\|\mathbf{v}^{u}_k\|_{\mathbf{S}_1}^{2}+\|\mathbf{a}^{u}_k\|_{\mathbf{S}_2}^{2}\right)+\sum_{m=t}^{t+L-1}\left(\|\mathbf{v}^{\mathrm{hand},+}_m\|_{\mathbf{V}}^{2}+\|\mathbf{a}^{\mathrm{hand}}_m\|_{\mathbf{A}}^{2}\right), \\
\mathbf{v}^{u}_k &= \frac{\mathbf{U}_k-\mathbf{U}_{k-1}}{\Delta_k}, \\
\mathbf{a}^{u}_k &= \frac{2(\mathbf{v}^{u}_k-\mathbf{v}^{u}_{k-1})}{\Delta_k+\Delta_{k-1}}.
\end{aligned}
\end{equation}
Here, $K$ is the number of control knots, $\Delta_k$ is the corresponding time interval, and executed-control history supplies the boundary values. The quantities $\mathbf{v}^{\mathrm{hand},+}_m$ and $\mathbf{a}^{\mathrm{hand}}_m$ are the post-step hand velocity and the hand acceleration returned by simulation step $m$. Reference velocities are not tracked, and no absolute control-energy penalty is used. We minimize Eq.~(\ref{eq:forceaware_full}) using parallel Adam optimization from one warm start and $N-1$ perturbed candidates. Each candidate retains its best finite iterate and is evaluated with the execution contact model; the lowest-cost candidate without numerical or simulator failures is selected. Contact forces are computed within the dynamics, and simulator parameters remain fixed during each optimization.

\section{Dataset selection}
\label{app:dataset-selection}
We select 50 right-hand grasping sequences from DexYCB and 80 right-hand in-hand manipulation sequences from HOT3D. For HOT3D, we extract clips with consecutive annotations for both the right-hand MANO pose and the target object pose. We retain only clips longer than $1\,\mathrm{s}$ in which the right hand contacts only the target object and the left hand is uninvolved throughout.

For ablation and system identification, we use five right-hand sequences from DexYCB subject~01: 20200709\_142802 (sugar\_box), 20200709\_144357 (gelatin\_box), 20200709\_145401 (banana), 20200709\_150613 (bowl), and 20200709\_153322 (foam\_brick). Each contains a single grasp-and-lift interaction and is used in full. We also select five HOT3D sequences: P0002\_5d79f6b9 (dumbbell\_5lb), P0003\_90feb35f (whiteboard\_eraser), P0003\_8243ea4a (food\_waffles), P0003\_ebdc6ff7 (can\_soup), and P0015\_b0c5102b (birdhouse\_toy), extracting clips using the same criteria above.

\section{Ablation Study Details}
\label{app:ablation}
Starting from the full method, we first replace reconstructed hand--object trajectories with dataset annotations while keeping the Gaussian object models unchanged. We then disable contact refinement by skipping the joint optimization of contact anchors and key robot configurations. The raw anchors directly guide joint kinematic optimization, initialized by fitting robot keypoints to MANO keypoints. Finally, disabling contact guidance removes stable-contact extraction, contact refinement, and contact-guidance objectives from both retargeting stages. All variants retain joint kinematic optimization with collision constraints and subsequent dynamics optimization. We use the five DexYCB and five HOT3D sequences listed in Appendix~\ref{app:dataset-selection}, evaluating all variants across seven robotic hands in ComFree-GS. All variants share optimization hyperparameters and use the success criteria defined in Sec.~\ref{sec:Experimental Setup}. For each dataset and hand, we compute the success rate over the five sequences for each seed and report the average across five seeds in Table~\ref{tab:retargeting-ablation}.

\section{Dynamics Retargeting Details and Additional Results}
\label{app:Dynamics Retargeting Results}
Both methods use warm-started MPC and interpolate control nodes to produce commands at $100\,\mathrm{Hz}$. The horizon $H$ is measured in control intervals, so the prediction-window duration equals $H$ times the control-node spacing. SPIDER retains its default configuration: a control-node spacing of $0.4\,\mathrm{s}$, $H=4$, a parallel batch size of 1024, and at most 16 optimization iterations. Our method uses a control-node spacing of $0.06\,\mathrm{s}$, $H=5$, a parallel batch size of 64, and at most 40 iterations. The resulting prediction windows are $1.6\,\mathrm{s}$ and $0.3\,\mathrm{s}$, respectively. We retain method-specific temporal configurations because imposing the same prediction window can disadvantage either optimizer. 

\begin{figure}[t]
\centering
\begin{subfigure}[t]{0.48\linewidth}
  \centering
  \includegraphics[width=\linewidth]{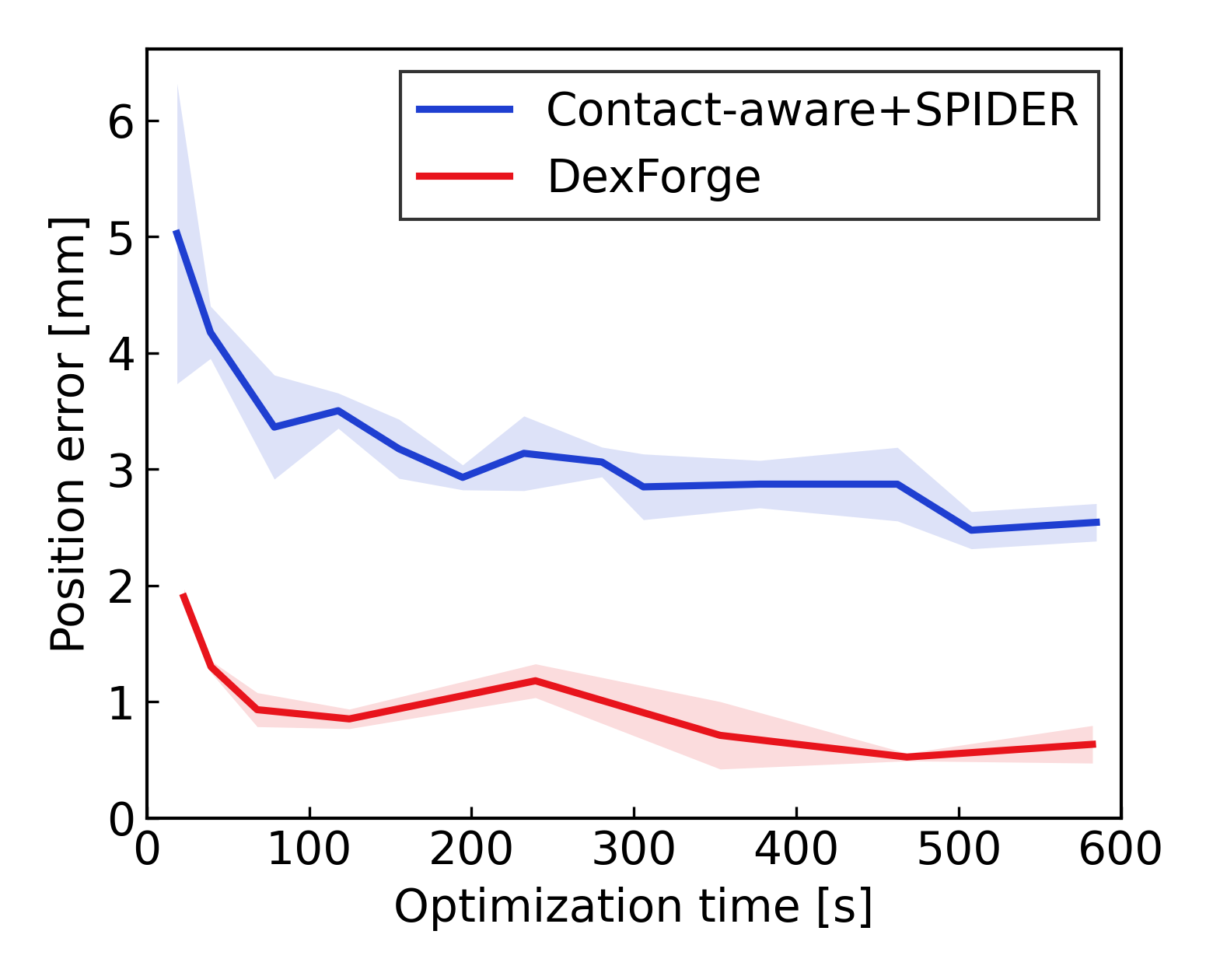}
  \caption{Object position error}
  \label{fig:runtime-position}
\end{subfigure}
\hfill
\begin{subfigure}[t]{0.48\linewidth}
  \centering
  \includegraphics[width=\linewidth]{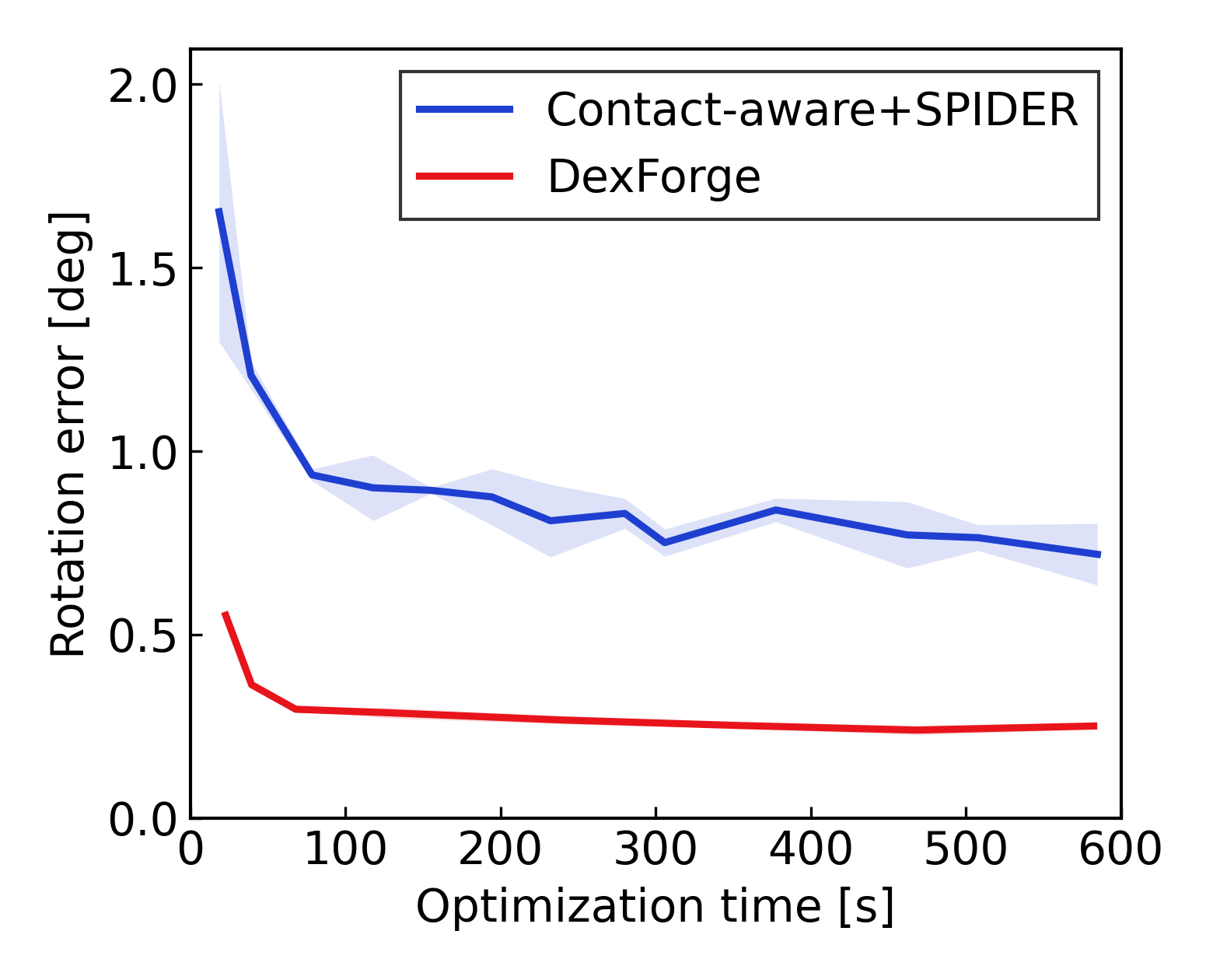}
  \caption{Object rotation error}
  \label{fig:runtime-rotation}
\end{subfigure}
\caption{Performance--runtime comparison between DexForge and Contact-aware+SPIDER. Object position (a) and
rotation (b) tracking errors are plotted against optimization time, with runtime varied through the optimization
iteration limit of each method. Curves show the mean over three runs and shaded regions indicate the standard
deviation. Under the same runtime budget, DexForge consistently achieves lower tracking errors.}
\label{fig:runtime-comparison}
\end{figure}

To examine the performance--runtime trade-off, we compare DexForge and Contact-aware+SPIDER on a fixed subset of ablation sequences successfully retargeted by both methods across all tested budgets. After object drops, smaller tracking errors do not necessarily indicate better retargeting; failures are instead reflected in the success rates reported above. For each method, we vary only the optimization iteration limit and measure total optimization time, disabling SPIDER's early stopping to control the budget through the iteration limit. All other MPC settings remain at their respective defaults. Both methods use identical kinematic references, stable-contact guidance, and simulation environment (ComFree-GS), running sequentially on a single RTX 4090 D. Each setting is repeated three times. Figure~\ref{fig:runtime-comparison} reports the mean and standard deviation of object position and rotation tracking errors at different runtimes. On this shared successful subset, DexForge achieves lower tracking errors under the same runtime budget, demonstrating more precise trajectory refinement.

\begin{figure}[t]
\centering
\begin{subfigure}[t]{0.48\linewidth}
  \centering
  \includegraphics[width=\linewidth]{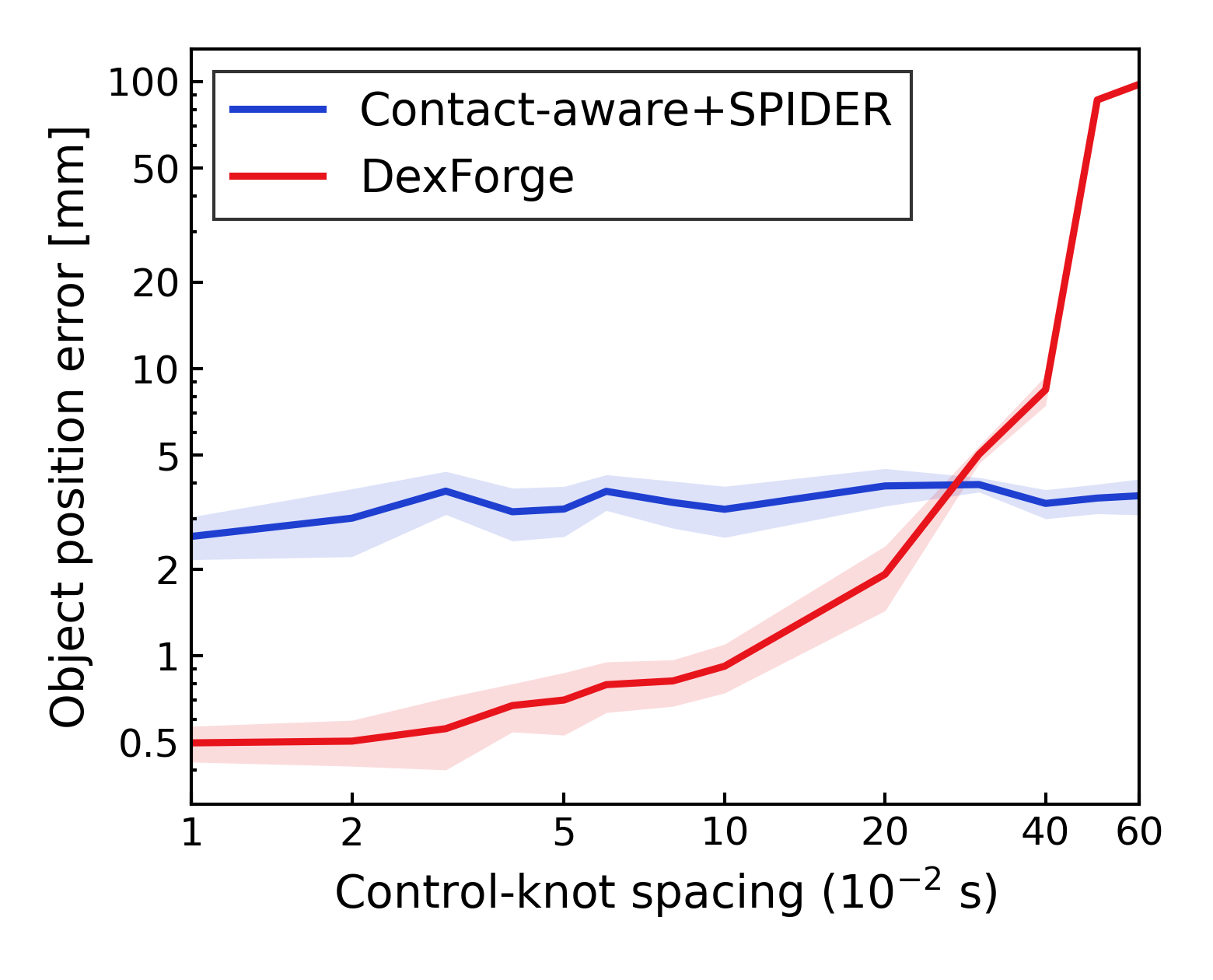}
  \caption{Object position error}
\end{subfigure}
\hfill
\begin{subfigure}[t]{0.48\linewidth}
  \centering
  \includegraphics[width=\linewidth]{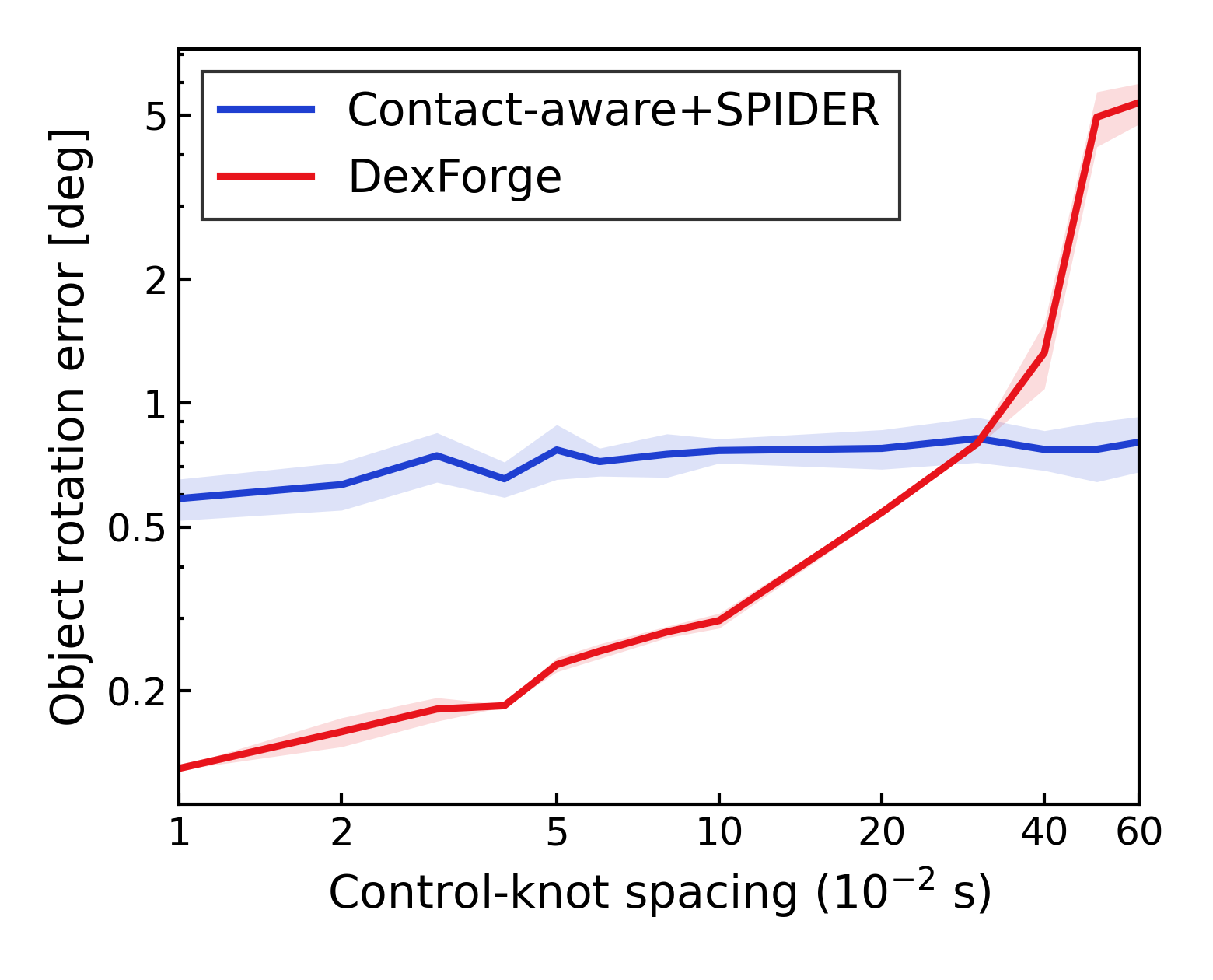}
  \caption{Object rotation error}
\end{subfigure}
\caption{Sensitivity to control-knot spacing under a matched $200\,\mathrm{s}$ optimization budget and $H=4$ control intervals. Errors are computed on a fixed subset of ablation sequences successfully retargeted by both methods under their default configurations; at other spacings, all runs on this subset are included regardless of success. Panels show object position (a) and rotation (b) tracking errors. Curves show mean errors over $3$ runs, with shaded regions indicating $\pm1$ standard deviation. Contact-aware+SPIDER is largely insensitive to knot spacing, whereas DexForge achieves lower errors at finer spacings but degrades as the spacing and prediction-window duration increase, consistent with greater difficulty in gradient propagation through longer rollouts.}
\label{fig:knot-spacing}
\end{figure}

We further examine sensitivity to control-knot spacing by comparing DexForge and Contact-aware+SPIDER with $H=4$ control intervals, choosing method-specific iteration limits at each spacing to obtain optimization runtimes close to $200\,\mathrm{s}$. We sweep the knot spacing over $\{0.01,0.02,\ldots,0.10,0.20,0.30,0.40,0.60\}\,\mathrm{s}$, varying the prediction-window duration accordingly while keeping the interpolated command rate at $100\,\mathrm{Hz}$. As shown in Figure~\ref{fig:knot-spacing}, on sequences completed by both methods, the tracking errors of Contact-aware+SPIDER are largely insensitive to knot spacing, remaining within approximately $2.5$--$4\,\mathrm{mm}$ and $0.6$--$0.8^\circ$ across the sweep, so its default spacing does not disadvantage its tracking accuracy. In contrast, DexForge achieves substantially lower errors at fine spacings, but its advantage diminishes and eventually reverses beyond approximately $0.3\,\mathrm{s}$, consistent with greater difficulty in gradient-based optimization over longer rollouts. This motivates the fine knot spacing used by DexForge.

This analysis, however, only covers sequences that both methods already complete and does not show whether finer spacing could reduce SPIDER's failures on HOT3D, where in-hand manipulation may involve higher-frequency motions. We therefore evaluate Contact-aware+SPIDER with $0.01\,\mathrm{s}$ knot spacing on the full HOT3D set. Its success rate increases only marginally, from $0.6\%$ to $6.0 \pm 3.7\%$, remaining far below that of DexForge ($34.9 \pm 2.1\%$). Moreover, even with accurate physical parameters, its open-loop transfer success to MuJoCo-CPU on the ablation subset is only $3.1 \pm 2.9\%$, below the $15.4 \pm 2.6\%$ achieved with the default spacing after identification (Table~\ref{tab:sim-to-sim-transfer}). This suggests that the resulting high-frequency controls overfit the source simulator and reduce transfer robustness. DexForge mitigates this issue through its coarser $0.06\,\mathrm{s}$ spacing and explicit control-smoothness regularization (Eq.~\ref{eq:forceaware_smooth}). We therefore retain each method's default temporal configuration in the main evaluation.

\section{Sim-to-Sim Transfer Details}
\label{app:sim-to-sim-transfer}
We evaluate open-loop transfer from ComFree-GS to MuJoCo-CPU, where target scenes use mesh-based object models with known or estimated physical properties. In ComFree-GS, all objects are initialized with a mass of $0.2\,\mathrm{kg}$, a contact friction coefficient of 1.0, a distance offset of zero, and an isotropic inertia of $0.001\,\mathrm{kg\,m^2}$. Although the formulation in Section~\ref{sec:system identification} supports identifying the full parameter set $\theta$, in these experiments we identify only the mass, friction coefficient, and distance offset, keeping the inertia fixed. For each sequence, we first optimize a complete retargeted control trajectory in ComFree-GS and replay a contact-rich segment in MuJoCo-CPU to collect target observations. Parameters are identified separately for each sequence by minimizing discrepancies between the simulated and recorded target trajectories under the same fixed controls. All replays use target-position commands at $100\,\mathrm{Hz}$ with matched initial states. Gradient-only identification uses Adam for 100 iterations with a learning rate of $10^{-3}$ and a gradient-clipping threshold of 10. Sampling-only identification uses CEM for 100 iterations with a population of 100, an elite fraction of $10\%$, an initial standard deviation of 1.0 in the raw parameter space, a minimum standard deviation of 0.01, and a smoothing coefficient of 0.7. TSMC uses 64 particles, a prior standard deviation of 0.5, a temperature parameter of 1.0, an ESS ratio of 0.7, and at most eight tempering stages with systematic resampling. Each stage performs four MALA steps with a step size of $10^{-4}$. Robust filtering with a threshold of $z=3.5$ is then applied before constructing an energy-weighted parameter distribution. After identification, we update the source parameters and rerun dynamics retargeting in ComFree-GS, then replay the resulting control trajectories open-loop in MuJoCo-CPU. Evaluation uses the same subsets as the ablation study across seven robotic hands and five seeds, with the transfer success criteria specified in the main text.

\section{Real-World Open-Loop Replay Details and Results}
\label{app:real-replay}
We replay optimized wrist and finger references on a Unitree G1 equipped with a LEAP Hand v1. The hand is mounted on the right wrist using a custom adapter, whose wrist-to-hand transformation is incorporated into the forward-kinematic model. Before replay, we place the object at the origin of a table-mounted motion-capture rigid-body frame with a predefined orientation to establish its initial pose. We then transform the precomputed wrist trajectory accordingly, select a reachable starting point, and move the robot to the corresponding configuration. A $100\,\mathrm{Hz}$ process uses Unitree's arm IK solver to generate arm joint-position targets and sends them together with the precomputed finger-position targets. Motion-capture torso poses compensate for body motion to preserve the intended world-frame wrist trajectory. A separate $50\,\mathrm{Hz}$ process runs NVIDIA's Decoupled WBC to maintain whole-body balance. Replay is open-loop with respect to object interaction: the precomputed manipulation references are not updated using object-state feedback, while torso compensation and balance control remain feedback-driven.

\paragraph{Results.}
We conduct 10 trials for each of the five objects, resetting the object to its predefined initial pose before each trial. Since object poses are not tracked during execution, a trial is considered successful if the object is stably grasped and held for $5\,\mathrm{s}$ without dropping. This criterion measures executability rather than tracking accuracy. The success counts are 10/10 for the gelatin box, 8/10 for the pudding box, 6/10 for the wood block, 10/10 for the power drill, and 6/10 for the tomato soup can, yielding an overall success rate of 80\% (40/50). Videos are available on the project page.

\end{document}